\documentclass[acmsmall]{acmart}
\AtBeginDocument{%
  }

\setcopyright{acmlicensed}
\copyrightyear{2018}
\acmYear{2018}
\acmDOI{XXXXXXX.XXXXXXX}

\acmJournal{JACM}
\acmVolume{37}
\acmNumber{4}
\acmArticle{111}
\acmMonth{8}

\usepackage{rotating}
\usepackage{makecell}
\usepackage{multirow}
\usepackage{colortbl}
\usepackage{tikz} 
\usepackage{fontawesome5}
\begin{document}


\title{Review of Hallucination Understanding in Large Language and Vision Models} 

\author{Ho Zheng Yi}
\email{zhengyi001@e.ntu.edu.sg}
\affiliation{%
  \institution{Nanyang Technological University}
  \country{Singapore}
}
\author{Liang Siyuan}
\email{pandaliang521@gmail.com}
\affiliation{%
  \institution{Nanyang Technological University}
  \country{Singapore}
}
\author{Tao Dacheng}
\email{dacheng.tao@e.ntu.edu.sg}
\affiliation{%
  \institution{Nanyang Technological University}
  \country{Singapore}
}

\renewcommand{\shortauthors}{Zhengyi Ho, Siyuan Liang, Dacheng Tao}


\begin{abstract}
The widespread adoption of large language and vision models in real-world applications has made urgent the need to address hallucinations—instances where models produce incorrect or nonsensical outputs. These errors can propagate misinformation during deployment, leading to both financial and operational harm. Although much research has been devoted to mitigating hallucinations, our understanding of it is still incomplete and fragmented. Without a coherent understanding of hallucinations, proposed solutions risk mitigating surface symptoms rather than underlying causes, limiting their effectiveness and generalizability in deployment. To tackle this gap, we first present a unified, multi-level framework for characterizing both image and text hallucinations across diverse applications, aiming to reduce conceptual fragmentation. We then link these hallucinations to specific mechanisms within a model’s lifecycle, using a task-modality interleaved approach to promote a more integrated understanding. Our investigations reveal that hallucinations often stem from predictable patterns in data distributions and inherited biases. By deepening our understanding, this survey provides a foundation for developing more robust and effective solutions to hallucinations in real-world generative AI systems.
\end{abstract}




\maketitle

\section{Introduction}
\label{sec:introduction}
Large Language Models (LLMs), Large Vision-Language Models (LVLMs), and Text-to-Vision Models (TVMs) now power numerous real-world applications that impact millions of users. As of 2024, over 77,000 organisations use GitHub's Copilot LLM for software development \citep{copilot-usage}. In the professional media sector, Adobe's Firefly TVM has surpassed 4.5 billion generations \citep{adobe-firefly-usage}. ChatGPT now supports over 1 million enterprise users \citep{chatgpt-usage} for daily tasks, while Google’s Gemini LVLM is automating complex image-text workflows across industries \citep{gemini-usage}. Despite their widespread adoption, these models often generate incorrect, inconsistent, or incoherent content---a phenomenon known as hallucinations \citep{tvm-hallucination-study,code-hallucination-study,lvlm-hallucination-study,chatgpt-hallucination-study}. Hallucinations can cause tangible harm: flawed code suggestions compromise software reliability \citep{code-incorrect-harm}, incoherent AI-generated media reduces viewer engagement \citep{media-incoherent-harm}, and inconsistent image-text analytics degrades workflow quality \citep{image-text-inconsistent-harm}. As organisations increasingly rely on these models to shape downstream products and services used by millions, addressing hallucinations in language and vision models is now a critical challenge.

Despite extensive efforts to address hallucinations, two critical gaps remain. First, our understanding of hallucinations remains limited. Most current research largely focus on mitigation strategies, often developed in response to observed failure cases, rather than grounded in a principled understanding of why hallucinations occur. As a result, many mitigation techniques may remain reactive and incomplete, which hinders their effectiveness and performance. Second, research on hallucinations remains fragmented and insufficiently systematised. Different studies frequently adopt narrow or anecdotal definitions tailored to specific tasks or modalities. This makes it difficult to draw broader insights or identify shared failure patterns that may reveal deeper underlying causes. As a result, the development of more generalisable and robust mitigation strategies may be significantly impeded. Addressing these two gaps in understanding and fragmentation is essential for reducing the risks of harm posed by hallucinations in real-world applications.


\begin{figure}
    \centering
    \begin{tikzpicture}
        \node[anchor=south west, inner sep=0] (image) at (0,0) {\includegraphics[width=\textwidth]{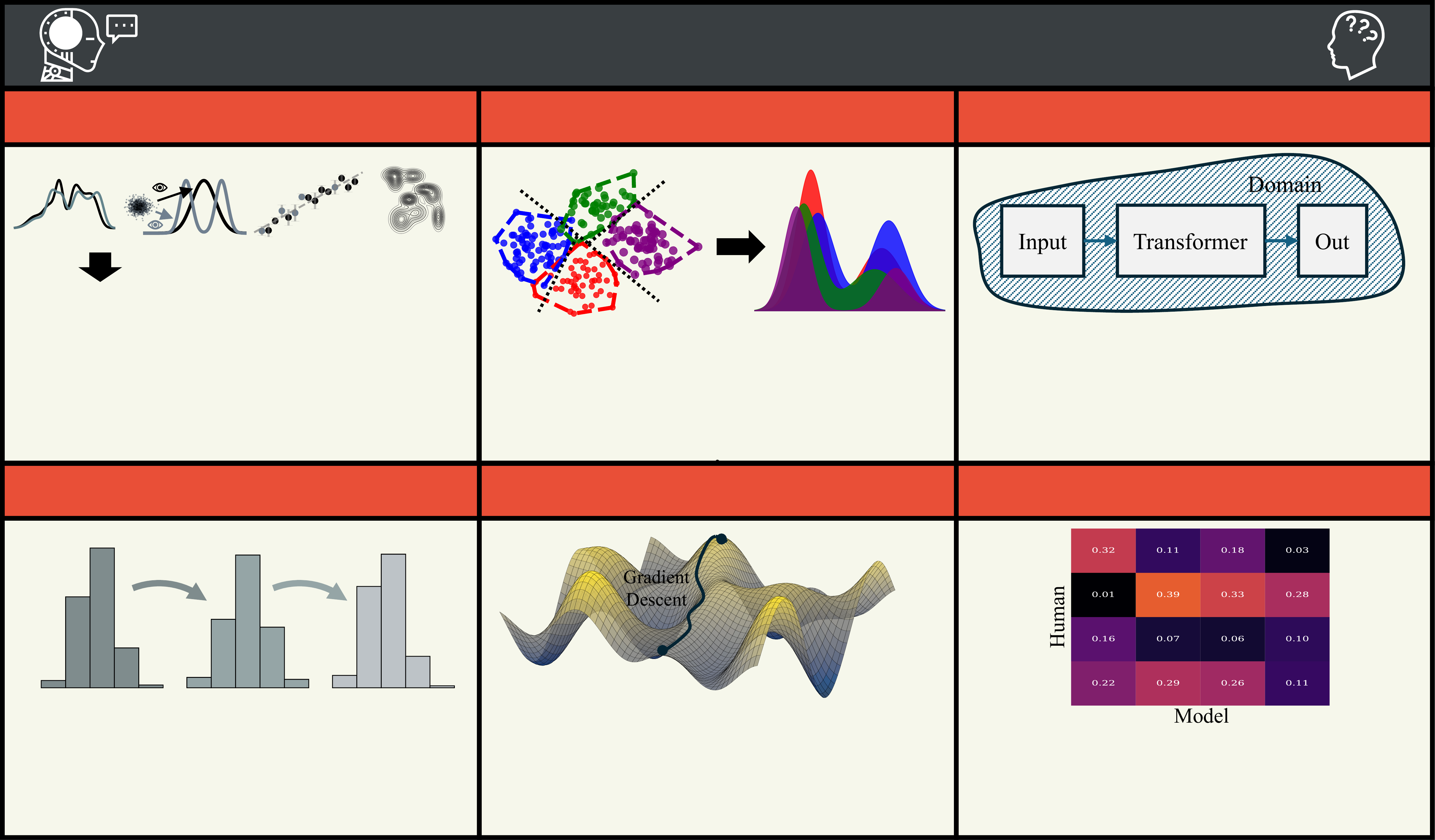}};

        \begin{scope}[x={(image.south east)}, y={(image.north west)}]

        \node[anchor=center,font=\color{white}] at (0.5,0.95) {\large{Understanding Hallucinations in Large visual and Language Models}};

        \node[anchor=center] at (0.165,0.8625) {\small{\S\ref{subsec:hallucination-definitions} Hallucination Definition}};
        \node[anchor=center] at (0.5,0.8625) {\small{\S\ref{subsec:training-data-factors} Training Data Factors}};
        \node[anchor=center] at (0.835,0.8625) {\small{\S\ref{subsec:architectural-limitations} Architectural Limitations}};

        \node[anchor=center] at (0.18,0.6825) {\footnotesize{MOWI Framework}};
        \node[anchor=west] at (0,0.625) {\footnotesize{\S\ref{subsubsec:model-level-halludef} Model}};
        \node[anchor=west] at (0,0.575) {\footnotesize{\S\ref{subsubsec:observer-level-halludef} Observer}};
        \node[anchor=west] at (0,0.525) {\footnotesize{\S\ref{subsubsec:world-level-halludef} World}};
        \node[anchor=west] at (0,0.475) {\footnotesize{\S\ref{subsubsec:input-level-halludef} Input}};

        \node[anchor=west] at (0.135,0.625) {\footnotesize{$d\bigl(P_{\theta}(x|c),P_{\text{real}}(x|c)\bigr)$}};
        \node[anchor=west] at (0.135,0.575) {\footnotesize{$d\bigl(P_{\theta}(x|c),P_{O_{i}}(x|c)\bigr)$}};
        \node[anchor=west] at (0.1348,0.525) {\footnotesize{$\bigl(U_{\text{ep}}+U_{\text{al}}\bigr)$}};
        \node[anchor=west] at (0.1348,0.475) {\footnotesize{$\bigl(g(c,\mathcal{X}_{\text{train}})+\nu(c)\bigr)$}};

        \node[anchor=west] at (0.332,0.592) {\footnotesize{\S\ref{subsubsec:salience-and-coverage} Salience and Coverage}};
        \node[anchor=west] at (0.332,0.552) {\footnotesize{\S\ref{subsubsec:memorisation} Memorisation}};
        \node[anchor=west] at (0.332,0.512) {\footnotesize{\S\ref{subsubsec:self-consumption} Self-Consumption}};
        \node[anchor=west] at (0.332,0.472) {\footnotesize{\S\ref{subsubsec:directional-asymmetries} Directional Asymmetries}};

        \node[anchor=west] at (0.664,0.592) {\footnotesize{\S\ref{subsubsec:attention-glitches} Attention Glitches}};
        \node[anchor=west] at (0.664,0.552) {\footnotesize{\S\ref{subsubsec:autoregressive-constraints} Autoregressive Constraints}};
        \node[anchor=west] at (0.664,0.512) {\footnotesize{\S\ref{subsubsec:incorrect-positional-encoding} Incorrect Positional Encoding}};
        \node[anchor=west] at (0.664,0.472) {\footnotesize{\S\ref{subsubsec:inductive-biases} Inductive Biases}};

        \node[anchor=center] at (0.165,0.415) {\small{\S\ref{subsec:inference-mechanisms} Inference Mechanisms}};
        \node[anchor=center] at (0.5,0.415) {\small{\S\ref{subsec:loss-and-optimisation} Loss and Optimisation}};
        \node[anchor=center] at (0.835,0.415) {\small{\S\ref{subsec:misleading-evaluations} Misleading Evaluations}};

        \node[anchor=west] at (0,0.105) {\footnotesize{\S\ref{subsubsec:few-shot-prompting} Few-Shot Quality}};
        \node[anchor=west] at (0,0.065) {\footnotesize{\S\ref{subsubsec:multi-agent-debates} Multi-Agent Debates}};
        \node[anchor=west] at (0,0.025) {\footnotesize{\S\ref{subsubsec:exposure-bias} Exposure Bias}};

        \node[anchor=west] at (0.332,0.145) {\footnotesize{\S\ref{subsubsec:pretraining-dynamics} Pretraining Dynamics}};
        \node[anchor=west] at (0.332,0.105) {\footnotesize{\S\ref{subsubsec:post-training-vulnerabilities} Post-Training Vulnerabilities}};
        \node[anchor=west] at (0.332,0.065) {\footnotesize{\S\ref{subsubsec:shortcut-learning} Shortcut Learning}};
        \node[anchor=west] at (0.332,0.025) {\footnotesize{\S\ref{subsubsec:heterogeneous-preferences} Heterogeneous Preferences}};

        \node[anchor=west] at (0.664,0.105) {\footnotesize{\S\ref{subsubsec:metric-blind-spots} Metric Blind Spots}};
        \node[anchor=west] at (0.664,0.065) {\footnotesize{\S\ref{subsubsec:biased-judges} Biased Judges}};
        \node[anchor=west] at (0.664,0.025) {\footnotesize{\S\ref{subsubsec:test-contamination} Test Contamination}};

        \end{scope}
  
    \end{tikzpicture}
    \caption{\textbf{Overview of the paper}. The first two sections provide an overview of the topic and related works. Section \ref{sec:definitions} defines key terms related to hallucinations and the model types under discussion. Section \ref{sec:root-causes-and-mechanisms} offers an in-depth review of the root causes and mechanisms of hallucinations. The final three sections build on these foundations to distil key insights, evaluate their broader implications, and propose future directions.}
    \label{fig:paper-overview}
    \Description[Overview of Paper]{An overview of the paper presented in the form of a chart with several boxes. Each box contains the title of different sections in the paper, together with iconography.}
\end{figure}

\begin{figure}
    \centering
    \begin{tikzpicture}
        \node[anchor=south west, inner sep=0] (image) at (0,0) {\includegraphics[width=\textwidth]{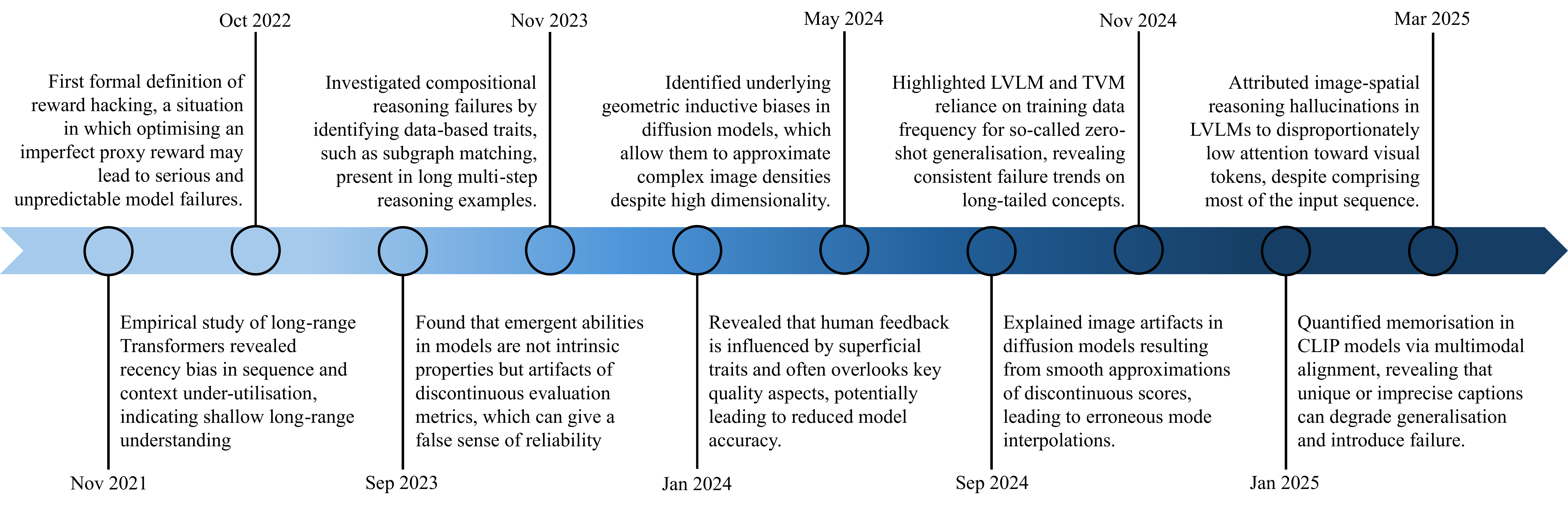}};

        \begin{scope}[x={(image.south east)}, y={(image.north west)}]

        \node[anchor=east] at (0.16575,0.887) {\scalebox{0.55}{\textbf{\citet{reward-hacking-rl}}}};
        \node[anchor=east] at (0.35,0.887) {\scalebox{0.55}{\textbf{\citet{faithandfate-llm}}}};
        \node[anchor=east] at (0.5385,0.887) {\scalebox{0.55}{\textbf{\citet{generalisation-diffusion-harmonic}}}};
        \node[anchor=east] at (0.727,0.887) {\scalebox{0.55}{\textbf{\citet{pretraining-concept-freq-multimodal}}}};
        \node[anchor=east] at (0.914,0.887) {\scalebox{0.55}{\textbf{\citet{spatial-reasoning-vlm-hard}}}};

        \node[anchor=west] at (0.068,0.415) {\scalebox{0.55}{\textbf{\citet{recency-bias}}}};
        \node[anchor=west] at (0.256,0.415) {\scalebox{0.55}{\textbf{\citet{emergent-mirage-llm}}}};
        \node[anchor=west] at (0.442,0.415) {\scalebox{0.55}{\textbf{\citet{human-feedback-not-gold-standard}}}};
        \node[anchor=west] at (0.6295,0.415) {\scalebox{0.55}{\textbf{\citet{diffusion-hallu-mode-interpolate}}}};
        \node[anchor=west] at (0.8165,0.415) {\scalebox{0.55}{\textbf{\citet{clip-mem-lvlm}}}};

        \end{scope}
  
    \end{tikzpicture}
    \caption{A timeline of representative works in the past four years exploring the understanding of failure modes in LLMs, LVLMs, and TVMs through a variety of methodological and theoretical lenses.}
    \label{fig:timeline-figure}
    \Description[Timeline Figure]{A timeline figure}
\end{figure}
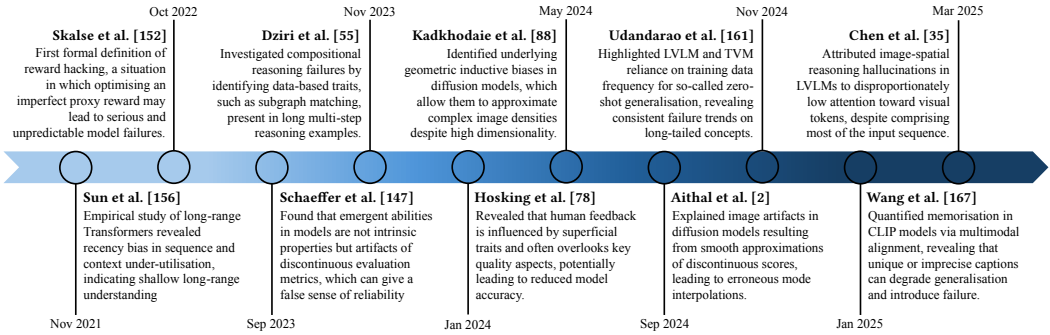

To address these gaps, we propose a detailed investigation and characterization of hallucinations. Our approach consists of three key contributions. First, we introduce a unified framework that offers a more general definition of hallucinations. In contrast to prior work, our framework accounts for modality- and task-specific differences, significantly improving its coverage and applicability. This helps reduce fragmentation in current discussions and promote a more cohesive discourse around hallucination phenomena. Second, we present a comprehensive survey of hallucination causes across LLMs, LVLMs, and TVMs. Our review is structured in a modality-interleaved fashion without rigid task delineation, allowing us to better identify shared failure patterns across systems. Crucially, we ground this survey in our proposed hallucination framework and trace its causes to identifiable mechanisms within a model’s lifecycle. This enables a deeper and more complete understanding of hallucinations. Finally, we consolidate insights from the survey to identify recurring themes and suggest future directions. An overview of the paper is provided in Figure \ref{fig:paper-overview}, with recent understanding efforts showcased in Figure \ref{fig:timeline-figure}. By offering a unified definition and root cause review, we aim to improve hallucination understanding and support the development of more generalisable and effective solutions, thus reducing the risks posed by AI systems in real-world applications.

\section{Related Works}
\label{sec:related_works}
\citet{jiziwei-nlg-hallu-survey} surveyed language hallucinations and defined them as outputs contradicting or unverifiable against source content, with task-specific criteria. \citet{truthfulqa} explored language hallucination evaluation and defined truth as verifiable real-world claims, while treating debatable viewpoints as hallucinations. \citet{leihuang-llm-hallu-survey} reviewed LLM hallucination mitigation and evaluation, briefly discussed their origins, and defined them based on categories derived from task-dependent interpretations of anecdotal examples. \citet{baizechen-lvlm-hallu-survey} surveyed LVLM hallucination mitigation and evaluation, briefly discussed their origins, and defined them using task-specific examples. \citet{sahoo-text-image-video-audio-hallu-survey} surveyed multimodal hallucination mitigation and evaluation, defining them with text-dependent, task-specific examples. \citet{kamali-tvm-hallu-survey} focused on LVM hallucination evaluation and defined them based on collated anecdotal observations. 

In contrast, our survey comprehensively reviews the origins of hallucinations by tracing their causes and mechanisms throughout a model’s life cycle. We move away from evaluation and mitigation to focus on uncovering root causes. Additionally, we extend our discussions beyond LLMs to include LVLMs and TVMs. Rather than focusing solely on either language or image, we identify both shared and unique hallucination patterns across modalities. Finally, we adopt a general and systematic definition of hallucinations. We move away from anecdotal and task-specific definitions of hallucinations, relying on a more structured and formal framework.

\section{Definitions}
\label{sec:definitions}
\subsection{Hallucination Definition}
\label{subsec:hallucination-definitions}

To derive a general definition of hallucinations, we first construct a framework with four levels: model, observer, world, and input (MOWI). Each of these four levels outlines root mechanisms by which a model's learned distribution fails to produce satisfactory and grounded outputs.

\subsubsection{Model Level} 
\label{subsubsec:model-level-halludef}
Density estimation errors occur when the learned distribution \( P_{\theta}(x|c) \) diverges from the true data distribution \( P_{\text{real}}(x|c) \). Although models are high-dimensional, real-world data typically lies on a lower-dimensional manifold due to inherent structures \citep{manifold-hypothesis-1,manifold-hypothesis-2}. If a training dataset \( X = \{x_i\}_{i=1}^{N} \) sampled from \( P_{\text{real}}(x|c) \) lies on a \( d\)-dimensional manifold, a model approximates \( P_{\text{real}}(x|c) \) through a parametric function \( p_{\theta}: \mathbb{R}^d \to \mathbb{R} \) \citep{statistical-learning-textbook}. This learnt approximation can yield both interpolation errors, where the model samples within high‐probability regions of the data manifold without capturing detailed variations, and extrapolation errors, where the model samples from regions with no real density such that \( P_{\theta}(x|c) \gg P_{\text{real}}(x|c) \). 

\subsubsection{Observer Level} 
\label{subsubsec:observer-level-halludef}
Belief variations occur when an output \(\displaystyle x\sim P_{\theta}(x | c)\) diverges from an observer’s viewpoint. Each observer’s  \(O_i\) viewpoint \(\displaystyle P_{O_i}(x|c)\) reflects individual epistemic frameworks. This scrutiny is trivial when \(\displaystyle P_{O_i}(x|c)\) overlaps significantly across observers, such as in common facts. More interesting is when \(\displaystyle P_{O_i}(x|c)\) varies greatly, such as in emerging scientific terminologies (e.g. the exact definition of a LLM), debatable medical procedures (e.g. pulpotomies for irreversible pulpitis) \citep{debatable-medical-procedures}, or ambiguous image scenes. Here, multiple plausible sources with equally verifiable and scientific evidence exist. Although typically framed as a lack of helpfulness in academic settings, real-world users often perceive such outputs as hallucinations: unfaithful, untrue or nonsensical outputs \citep{hci-user-intent-llm-1,hci-user-intent-llm-2}.

\subsubsection{World Level} 
\label{subsubsec:world-level-halludef}
Epistemic uncertainty arises when the model lacks data to accurately approximate \( P_{\text{real}}(x|c, t)\) at time \( t \), leading to high variance in \( P_{\theta}(x|c, t) \). This is described by the posterior variance over model parameters: \(U_{\text{ep}} = \mathbb{E}_{P(D)} [ \text{Var}(P_{\theta}(x|c, t) | D) ]\), where \( D \) is the training data. \(U_{\text{ep}}\) can be due to practical limits imposed by inaccessible knowledge, such as in esoteric, classified, or time-sensitive topics \citep{aleatoric-epistemic-uncertainty-ml}. Aleatoric uncertainty stems from the inherent randomness in \( P_{\text{real}}(x|c, t) \) itself, expressed as the irreducible variance in the true data distribution: \(U_{\text{al}} = \text{Var}(P_{\text{real}}(x|c, t))\). 

\subsubsection{Input Level} 
\label{subsubsec:input-level-halludef}
At the input level, hallucinations arise when the conditioning variable \( c \) in \( P_{\theta}(x | c) \) is sparse, contradictory, or out-of-distribution, forcing the model to operate beyond its learned priors. An input distribution \( c^* \sim P_{\text{real}}(c) \) that lies outside training support \( P_{\text{train}}(c^*) \approx 0 \) results in high-entropy and unreliable outputs \(H(P_{\theta}(x | c^*)) = - \sum_{x} P_{\theta}(x | c^*) \log P_{\theta}(x | c^*)\), compared to well-conditioned cases. This issue is exacerbated in interactive settings, such as open-ended dialogue and multi-agent systems, where the \(c\) itself evolves based on prior outputs.

\subsubsection{General Definition} 
Putting the four framework levels (abbreviated as MOWI) together, a general definition of hallucinations \(P_\mathrm{Hal}(x|c,O_i)\) can now be derived: 

\begin{equation*}
    P_\mathrm{Hal}(x|c,O_i)=\Phi\Bigl[\underbrace{d\Bigl(P_{\theta}(x|c),P_{\text{real}}(x|c)\Bigr)}_{\text{Model}},\;\underbrace{d\Bigl(P_{\theta}(x|c),P_{O_{i}}(x|c)\Bigr)}_{\text{Observer}},\;\underbrace{\Bigl(U_{\text{ep}}+U_{\text{al}}\Bigr)}_{\text{World}},\;\underbrace{\Bigl(g(c,\mathcal{X}_{\text{train}})+\nu(c)\Bigr)}_{\text{Input}}\Bigr].
\end{equation*}
\(P_\mathrm{Hal}(x|c,O_i)\) is the probability of a hallucination, \(\Phi\) is monotonic non-decreasing in each argument, \(d\) is some distance function, \(g(c,\mathcal{X}_{\text{train}})\) measures how far \(c\) is from the training‐data manifold, and \(\nu(c)\) is a function that measures sparsity and contradictions in \(c\). To demonstrate the general applicability of our hallucination definition, we apply it to challenging scenarios across modalities and contrast it against existing definitions in Table \ref{tab:hallucination-definition}.

\begin{table}
\centering
\caption{Examples of our hallucination definition applied to broad scenarios. A \textbf{\textcolor{red}{$\times$}} indicates the case does \textbf{not} qualify as a hallucination by definition, while a \textbf{\textcolor[rgb]{0,0.502,0}{$\checkmark$}} means it is. A "-" means the definition is not applicable. Our definition covers a broad range of cases across tasks and modalities in a unified manner.}
\label{tab:hallucination-definition}
\arrayrulecolor[rgb]{0.502,0.502,0.502}
\resizebox{\linewidth}{!}{%
\begin{tabular}{!{\color{black}\vrule}c!{\color{black}\vrule}l!{\color{black}\vrule}l!{\color{black}\vrule}c|c|c|c!{\color{black}\vrule}c|c|c|c!{\color{black}\vrule}} 
\arrayrulecolor{black}\hline
\multicolumn{1}{!{\color{black}\vrule}l}{} & \multicolumn{1}{l}{} &  & \multicolumn{4}{c!{\color{black}\vrule}}{\begin{tabular}[c]{@{}c@{}}Existing\\Definitions\end{tabular}} & \multicolumn{4}{c!{\color{black}\vrule}}{\begin{tabular}[c]{@{}c@{}}Our\\Definition\end{tabular}} \\ 
\cline{4-11}
\multicolumn{1}{!{\color{black}\vrule}l}{} & \multicolumn{1}{l}{Task} & Scenarios & \multicolumn{1}{l!{\color{black}\vrule}}{\citep{leihuang-llm-hallu-survey}} & \multicolumn{1}{c!{\color{black}\vrule}}{\citep{jiziwei-nlg-hallu-survey}} & \multicolumn{1}{c!{\color{black}\vrule}}{\citep{baizechen-lvlm-hallu-survey}} & \multicolumn{1}{c!{\color{black}\vrule}}{\citep{kamali-tvm-hallu-survey}} & \multicolumn{1}{c!{\color{black}\vrule}}{M} & \multicolumn{1}{c!{\color{black}\vrule}}{O} & \multicolumn{1}{c!{\color{black}\vrule}}{W} & \multicolumn{1}{c!{\color{black}\vrule}}{I} \\ 
\hline
\multirow{6}{*}{\rotatebox[origin=c]{90}{LLMs}} & \multirow{2}{*}{Coding} & Code is correct, but relies on deprecated & \multirow{2}{*}{\textbf{\textcolor{red}{$\times$}}} & \multirow{2}{*}{\textbf{\textbf{\textcolor{red}{$\times$}}}} & \multirow{2}{*}{-} & \multirow{2}{*}{-} & \multirow{2}{*}{\textbf{\textbf{\textcolor[rgb]{0,0.502,0}{$\checkmark$}}}} & \multirow{2}{*}{-} & \multirow{2}{*}{\textbf{\textbf{\textcolor[rgb]{0,0.502,0}{$\checkmark$}}}} & \multirow{2}{*}{-} \\
 &  & libraries and suboptimal algorithms.~ &  &  &  &  &  &  &  &  \\ 
\arrayrulecolor[rgb]{0.502,0.502,0.502}\cline{2-11}
 & \multirow{2}{*}{Summarisation} & Summary is faithful, but is extractive & \multirow{2}{*}{\textbf{\textbf{\textcolor{red}{$\times$}}}} & \multirow{2}{*}{\textbf{\textbf{\textcolor{red}{$\times$}}}} & \multirow{2}{*}{-} & \multirow{2}{*}{-} & \multirow{2}{*}{\textbf{\textbf{\textcolor[rgb]{0,0.502,0}{$\checkmark$}}}} & \multirow{2}{*}{\textbf{\textbf{\textcolor[rgb]{0,0.502,0}{$\checkmark$}}}} & \multirow{2}{*}{-} & \multirow{2}{*}{-} \\
 &  & and lacks abstractive condensation. &  &  &  &  &  &  &  &  \\ 
\cline{2-11}
 & \multirow{2}{*}{Generative QA} & Question relies on a false premise, contains & \multirow{2}{*}{\textbf{\textcolor[rgb]{0,0.502,0}{$\checkmark$}}} & \multirow{2}{*}{\textbf{\textbf{\textcolor{red}{$\times$}}}} & \multirow{2}{*}{-} & \multirow{2}{*}{-} & \multirow{2}{*}{-} & \multirow{2}{*}{-} & \multirow{2}{*}{-} & \multirow{2}{*}{\textbf{\textbf{\textcolor[rgb]{0,0.502,0}{$\checkmark$}}}} \\
 &  & a contradiction or is incoherent. &  &  &  &  &  &  &  &  \\ 
\arrayrulecolor{black}\hline
\multirow{6}{*}{\rotatebox[origin=c]{90}{LVLMs}} & \multirow{2}{*}{Captioning} & Described visual elements of a scientific & \multirow{2}{*}{-} & \multirow{2}{*}{-} & \multirow{2}{*}{\textbf{\textbf{\textcolor{red}{$\times$}}}} & \multirow{2}{*}{-} & \multirow{2}{*}{\textbf{\textbf{\textcolor[rgb]{0,0.502,0}{$\checkmark$}}}} & \multirow{2}{*}{\textbf{\textbf{\textcolor[rgb]{0,0.502,0}{$\checkmark$}}}} & \multirow{2}{*}{-} & \multirow{2}{*}{\textbf{\textbf{\textcolor[rgb]{0,0.502,0}{$\checkmark$}}}} \\
 &  & chart correctly, but lacks semantic insight. &  &  &  &  &  &  &  &  \\ 
\arrayrulecolor[rgb]{0.502,0.502,0.502}\cline{2-11}
 & \multirow{2}{*}{Visual QA} & Predicting object trajectories or future & \multirow{2}{*}{-} & \multirow{2}{*}{-} & \multirow{2}{*}{\textbf{\textbf{\textcolor{red}{$\times$}}}} & \multirow{2}{*}{-} & \multirow{2}{*}{-} & \multirow{2}{*}{-} & \multirow{2}{*}{\textbf{\textbf{\textcolor[rgb]{0,0.502,0}{$\checkmark$}}}} & \multirow{2}{*}{-} \\
 &  & actions in a still image. &  &  &  &  &  &  &  &  \\ 
\cline{2-11}
 & \multirow{2}{*}{Detection} & Counting apples and red balls in a scene & \multirow{2}{*}{-} & \multirow{2}{*}{-} & \multirow{2}{*}{\textbf{\textbf{\textcolor{red}{$\times$}}}} & \multirow{2}{*}{-} & \multirow{2}{*}{-} & \multirow{2}{*}{\textbf{\textbf{\textcolor[rgb]{0,0.502,0}{$\checkmark$}}}} & \multirow{2}{*}{\textbf{\textbf{\textcolor[rgb]{0,0.502,0}{$\checkmark$}}}} & \multirow{2}{*}{\textbf{\textbf{\textcolor[rgb]{0,0.502,0}{$\checkmark$}}}} \\
 &  & with occlusions and photometric artifacts. &  &  &  &  &  &  &  &  \\ 
\arrayrulecolor{black}\hline
\multirow{6}{*}{\rotatebox[origin=c]{90}{TVMs}} & \multirow{2}{*}{Design} & Drawing objects with uncanny anomalies~ & \multirow{2}{*}{-} & \multirow{2}{*}{-} & \multirow{2}{*}{-} & \multirow{2}{*}{\textbf{\textbf{\textcolor[rgb]{0,0.502,0}{$\checkmark$}}}} & \multirow{2}{*}{\textbf{\textbf{\textcolor[rgb]{0,0.502,0}{$\checkmark$}}}} & \multirow{2}{*}{-} & \multirow{2}{*}{-} & \multirow{2}{*}{-} \\
 &  & and slight proportion errors. &  &  &  &  &  &  &  &  \\ 
\arrayrulecolor[rgb]{0.502,0.502,0.502}\cline{2-11}
 & \multirow{2}{*}{In-painting} & Losing structural and semantic consistency & \multirow{2}{*}{-} & \multirow{2}{*}{-} & \multirow{2}{*}{-} & \multirow{2}{*}{\textbf{\textbf{\textcolor{red}{$\times$}}}} & \multirow{2}{*}{-} & \multirow{2}{*}{\textbf{\textbf{\textcolor[rgb]{0,0.502,0}{$\checkmark$}}}} & \multirow{2}{*}{\textbf{\textbf{\textcolor[rgb]{0,0.502,0}{$\checkmark$}}}} & \multirow{2}{*}{\textbf{\textbf{\textcolor[rgb]{0,0.502,0}{$\checkmark$}}}} \\
 &  & within a local scene neighbourhood. &  &  &  &  &  &  &  &  \\ 
\cline{2-11}
 & \multirow{2}{*}{Generation} & Generating homogenised cityscape scenes & \multirow{2}{*}{-} & \multirow{2}{*}{-} & \multirow{2}{*}{-} & \multirow{2}{*}{\textbf{\textbf{\textcolor{red}{$\times$}}}} & \multirow{2}{*}{\textbf{\textbf{\textcolor[rgb]{0,0.502,0}{$\checkmark$}}}} & \multirow{2}{*}{\textbf{\textbf{\textcolor[rgb]{0,0.502,0}{$\checkmark$}}}} & \multirow{2}{*}{-} & \multirow{2}{*}{\textbf{\textbf{\textcolor[rgb]{0,0.502,0}{$\checkmark$}}}} \\
 &  & that do not reflect specific local styles. &  &  &  &  &  &  &  &  \\
\arrayrulecolor{black}\hline
\end{tabular}
}
\end{table}

\subsection{Model Definition}
We define and scope the three model types surveyed in this paper as follows. First, we refer to Large Language Models (LLMs) as transformer-based models pretrained on large-scale language corpora. These models may be further finetuned for preference alignment and task specialisation to perform text-to-text tasks, optionally with in-context learning. Second, we define Large Vision-Language Models (LVLMs) as transformer-based architectures comprising separately pretrained vision and language encoders, integrated via a multimodal fusion mechanism. These models may be subsequently finetuned and aligned on image-text datasets to perform text-and-image to text tasks, optionally with in-context learning. Third, we refer to Text-to-Image Vision Models (TVMs) as textually conditioned denoising diffusion models, typically using either transformer or convolutional architectures. These models may be finetuned to capture aesthetic preferences or stylistic qualities for text-to-image tasks. Having scoped the three model types discussed in this paper, we now examine the stages of their operational lifecycle and how each can introduce vulnerabilities that contribute to hallucinations. The process begins with pretraining, where models are exposed to large-scale datasets to develop broad foundational abilities. At this stage, issues related to the quality and distribution of training data (Section \ref{subsec:training-data-factors}) can play a major role in the early formation of hallucination tendencies. In addition to data, architectural design choices (Section \ref{subsec:architectural-limitations}) may encode limitations that affect a model’s capacity to learn generalisable patterns, often leading to persistent failure modes. Another important consideration is loss and optimisation behaviour during various training phases, including pretraining, alignment, and task-specific finetuning (Section \ref{subsec:loss-and-optimisation}). These dynamics affect the likelihood of hallucinations by influencing how models internalise patterns and respond to unseen inputs. Beyond training, evaluation practices (Section \ref{subsec:misleading-evaluations}) can significantly impact model quality. Vulnerabilities here may reinforce false signals of progress and allow hallucination-related issues to persist or worsen. Finally, inference brings its own set of risks. The way users interact with models and the degree to which inputs align with patterns learned during training (Section \ref{subsec:inference-mechanisms}) can exacerbate errors. These user-facing failures are often the most visible, carrying direct implications for reliability in real-world settings. The following sections discuss these factors in more detail, with an overview provided in Table \ref{tab:grand-table}.

\begin{table}
\centering
\caption{Summary of hallucination causes and mechanisms attributed to five distinct stages in a model's lifecycle. Each stage is further divided into 3-4 detailed categories. Each category traces specific hallucination types, as defined in Section \ref{subsec:hallucination-definitions}, to underlying causes rooted with identifiable mechanisms. The image \faCameraRetro\space and text \faFile*[regular] icons indicate the relevant hallucination modality, while the right-most column links to sections discussing the corresponding causes and mechanisms in greater detail.}
\label{tab:grand-table}
\resizebox{\linewidth}{!}{%
\begin{tabular}{|l|l|c|c|c|c|c|c|} 
\hline
\multicolumn{2}{|c|}{\multirow{2}{*}{\textbf{Root Causes and Mechanisms}}} & \multicolumn{4}{c|}{\textbf{Hallucinations}} & \multirow{2}{*}{\textbf{Modality}} & {\multirow{2}{*}{\textbf{Section}}} \\
\multicolumn{2}{|c|}{} & \multicolumn{1}{c}{\textbf{M}} & \multicolumn{1}{c}{\textbf{O}} & \multicolumn{1}{c}{\textbf{W}} & \multicolumn{1}{c|}{\textbf{I}} &  & \\ 
\hline\hline
\multirow{4}{*}{Training Data Factors} & Salience and Coverage & \raisebox{-0.33ex}{\textcolor[rgb]{0,0.502,0}{\faCheckCircle}} & \raisebox{-0.33ex}{\textcolor[rgb]{0.55,0.55,0.55}{\faCircle[regular]}} & \raisebox{-0.33ex}{\textcolor[rgb]{0,0.502,0}{\faCheckCircle}} & \raisebox{-0.33ex}{\textcolor[rgb]{0,0.502,0}{\faCheckCircle}} & \raisebox{-0.33ex}{\faCameraRetro\space\faFile*[regular]} & \ref{subsubsec:salience-and-coverage} \\
\cline{2-8}
 & Memorisation & \raisebox{-0.33ex}{\textcolor[rgb]{0,0.502,0}{\faCheckCircle}} & \raisebox{-0.33ex}{\textcolor[rgb]{0.55,0.55,0.55}{\faCircle[regular]}} & \raisebox{-0.33ex}{\textcolor[rgb]{0.55,0.55,0.55}{\faCircle[regular]}} & \raisebox{-0.33ex}{\textcolor[rgb]{0,0.502,0}{\faCheckCircle}} & \raisebox{-0.33ex}{\faCameraRetro\space\faFile*[regular]} & \ref{subsubsec:memorisation} \\
\cline{2-8}
 & Self-Consumption & \raisebox{-0.33ex}{\textcolor[rgb]{0,0.502,0}{\faCheckCircle}} & \raisebox{-0.33ex}{\textcolor[rgb]{0.55,0.55,0.55}{\faCircle[regular]}} & \raisebox{-0.33ex}{\textcolor[rgb]{0,0.502,0}{\faCheckCircle}} & \raisebox{-0.33ex}{\textcolor[rgb]{0.55,0.55,0.55}{\faCircle[regular]}} & \raisebox{-0.33ex}{\faCameraRetro\space\faFile*[regular]} & \ref{subsubsec:self-consumption} \\
\cline{2-8}
 & Directional Asymmetries & \raisebox{-0.33ex}{\textcolor[rgb]{0,0.502,0}{\faCheckCircle}} & \raisebox{-0.33ex}{\textcolor[rgb]{0.55,0.55,0.55}{\faCircle[regular]}} & \raisebox{-0.33ex}{\textcolor[rgb]{0.55,0.55,0.55}{\faCircle[regular]}} & \raisebox{-0.33ex}{\textcolor[rgb]{0.55,0.55,0.55}{\faCircle[regular]}} & \raisebox{-0.33ex}{\faCameraRetro\space\faFile*[regular]} & \ref{subsubsec:directional-asymmetries} \\
\hline\hline
\multirow{4}{*}{Achitectural Limitations} & Attention Glitches & \raisebox{-0.33ex}{\textcolor[rgb]{0,0.502,0}{\faCheckCircle}} & \raisebox{-0.33ex}{\textcolor[rgb]{0.55,0.55,0.55}{\faCircle[regular]}} & \raisebox{-0.33ex}{\textcolor[rgb]{0.55,0.55,0.55}{\faCircle[regular]}} & \raisebox{-0.33ex}{\textcolor[rgb]{0,0.502,0}{\faCheckCircle}} & \raisebox{-0.33ex}{\faCameraRetro\space\faFile*[regular]} & \ref{subsubsec:attention-glitches} \\
\cline{2-8}
 & Autoregressive Constraints & \raisebox{-0.33ex}{\textcolor[rgb]{0.55,0.55,0.55}{\faCircle[regular]}} & \raisebox{-0.33ex}{\textcolor[rgb]{0.55,0.55,0.55}{\faCircle[regular]}} & \raisebox{-0.33ex}{\textcolor[rgb]{0,0.502,0}{\faCheckCircle}} & \raisebox{-0.33ex}{\textcolor[rgb]{0,0.502,0}{\faCheckCircle}} & \raisebox{-0.33ex}{\faCameraRetro\space\faFile*[regular]} & \ref{subsubsec:autoregressive-constraints} \\
\cline{2-8}
 & Incorrect Positional Encoding & \raisebox{-0.33ex}{\textcolor[rgb]{0.55,0.55,0.55}{\faCircle[regular]}} & \raisebox{-0.33ex}{\textcolor[rgb]{0.55,0.55,0.55}{\faCircle[regular]}} & \raisebox{-0.33ex}{\textcolor[rgb]{0.55,0.55,0.55}{\faCircle[regular]}} & \raisebox{-0.33ex}{\textcolor[rgb]{0,0.502,0}{\faCheckCircle}} & \raisebox{-0.33ex}{\faFile*[regular]} & \ref{subsubsec:incorrect-positional-encoding} \\
\cline{2-8}
 & Inductive Biases & \raisebox{-0.33ex}{\textcolor[rgb]{0,0.502,0}{\faCheckCircle}} & \raisebox{-0.33ex}{\textcolor[rgb]{0.55,0.55,0.55}{\faCircle[regular]}} & \raisebox{-0.33ex}{\textcolor[rgb]{0,0.502,0}{\faCheckCircle}} & \raisebox{-0.33ex}{\textcolor[rgb]{0.55,0.55,0.55}{\faCircle[regular]}} & \raisebox{-0.33ex}{\faCameraRetro\space\faFile*[regular]} & \ref{subsubsec:inductive-biases} \\
\hline\hline
\multirow{3}{*}{Inference Mechanisms} & Few-Shot Quality & \raisebox{-0.33ex}{\textcolor[rgb]{0.55,0.55,0.55}{\faCircle[regular]}} & \raisebox{-0.33ex}{\textcolor[rgb]{0,0.502,0}{\faCheckCircle}} & \raisebox{-0.33ex}{\textcolor[rgb]{0.55,0.55,0.55}{\faCircle[regular]}} & \raisebox{-0.33ex}{\textcolor[rgb]{0,0.502,0}{\faCheckCircle}} & \raisebox{-0.33ex}{\faFile*[regular]} & \ref{subsubsec:few-shot-prompting} \\
\cline{2-8}
 & Multi-Agent Debates & \raisebox{-0.33ex}{\textcolor[rgb]{0.55,0.55,0.55}{\faCircle[regular]}} & \raisebox{-0.33ex}{\textcolor[rgb]{0.55,0.55,0.55}{\faCircle[regular]}} & \raisebox{-0.33ex}{\textcolor[rgb]{0.55,0.55,0.55}{\faCircle[regular]}} & \raisebox{-0.33ex}{\textcolor[rgb]{0,0.502,0}{\faCheckCircle}} & \raisebox{-0.33ex}{\faFile*[regular]} & \ref{subsubsec:multi-agent-debates} \\
\cline{2-8}
 & Exposure Bias & \raisebox{-0.33ex}{\textcolor[rgb]{0.55,0.55,0.55}{\faCircle[regular]}} & \raisebox{-0.33ex}{\textcolor[rgb]{0.55,0.55,0.55}{\faCircle[regular]}} & \raisebox{-0.33ex}{\textcolor[rgb]{0.55,0.55,0.55}{\faCircle[regular]}} & \raisebox{-0.33ex}{\textcolor[rgb]{0,0.502,0}{\faCheckCircle}} & \raisebox{-0.33ex}{\faCameraRetro\space\faFile*[regular]} & \ref{subsubsec:exposure-bias} \\
\hline\hline
\multirow{4}{*}{Loss and Optimisation} & Pretraining Dynamics & \raisebox{-0.33ex}{\textcolor[rgb]{0,0.502,0}{\faCheckCircle}} & \raisebox{-0.33ex}{\textcolor[rgb]{0.55,0.55,0.55}{\faCircle[regular]}} & \raisebox{-0.33ex}{\textcolor[rgb]{0,0.502,0}{\faCheckCircle}} & \raisebox{-0.33ex}{\textcolor[rgb]{0.55,0.55,0.55}{\faCircle[regular]}} & \raisebox{-0.33ex}{\faCameraRetro\space\faFile*[regular]} & \ref{subsubsec:pretraining-dynamics} \\
\cline{2-8}
 & Post-Training Vulnerabilities & \raisebox{-0.33ex}{\textcolor[rgb]{0.55,0.55,0.55}{\faCircle[regular]}} & \raisebox{-0.33ex}{\textcolor[rgb]{0,0.502,0}{\faCheckCircle}} & \raisebox{-0.33ex}{\textcolor[rgb]{0.55,0.55,0.55}{\faCircle[regular]}} & \raisebox{-0.33ex}{\textcolor[rgb]{0,0.502,0}{\faCheckCircle}} & \raisebox{-0.33ex}{\faFile*[regular]} & \ref{subsubsec:post-training-vulnerabilities} \\
\cline{2-8}
 & Shortcut Learning & \raisebox{-0.33ex}{\textcolor[rgb]{0,0.502,0}{\faCheckCircle}} & \raisebox{-0.33ex}{\textcolor[rgb]{0.55,0.55,0.55}{\faCircle[regular]}} & \raisebox{-0.33ex}{\textcolor[rgb]{0,0.502,0}{\faCheckCircle}} & \raisebox{-0.33ex}{\textcolor[rgb]{0.55,0.55,0.55}{\faCircle[regular]}} & \raisebox{-0.33ex}{\faCameraRetro\space\faFile*[regular]} & \ref{subsubsec:shortcut-learning} \\
\cline{2-8}
 & Heterogeneous Preferences & \raisebox{-0.33ex}{\textcolor[rgb]{0.55,0.55,0.55}{\faCircle[regular]}} & \raisebox{-0.33ex}{\textcolor[rgb]{0,0.502,0}{\faCheckCircle}} & \raisebox{-0.33ex}{\textcolor[rgb]{0.55,0.55,0.55}{\faCircle[regular]}} & \raisebox{-0.33ex}{\textcolor[rgb]{0.55,0.55,0.55}{\faCircle[regular]}} & \raisebox{-0.33ex}{\faCameraRetro\space\faFile*[regular]} & \ref{subsubsec:heterogeneous-preferences} \\
\hline\hline
\multirow{3}{*}{Misleading Evaluations} & Metric Blind Spots & \raisebox{-0.33ex}{\textcolor[rgb]{0,0.502,0}{\faCheckCircle}} & \raisebox{-0.33ex}{\textcolor[rgb]{0.55,0.55,0.55}{\faCircle[regular]}} & \raisebox{-0.33ex}{\textcolor[rgb]{0.55,0.55,0.55}{\faCircle[regular]}} & \raisebox{-0.33ex}{\textcolor[rgb]{0.55,0.55,0.55}{\faCircle[regular]}} & \raisebox{-0.33ex}{\faCameraRetro\space\faFile*[regular]} & \ref{subsubsec:metric-blind-spots} \\
\cline{2-8}
 & Biased Judges & \raisebox{-0.33ex}{\textcolor[rgb]{0,0.502,0}{\faCheckCircle}} & \raisebox{-0.33ex}{\textcolor[rgb]{0,0.502,0}{\faCheckCircle}} & \raisebox{-0.33ex}{\textcolor[rgb]{0.55,0.55,0.55}{\faCircle[regular]}} & \raisebox{-0.33ex}{\textcolor[rgb]{0,0.502,0}{\faCheckCircle}} & \raisebox{-0.33ex}{\faCameraRetro\space\faFile*[regular]} & \ref{subsubsec:biased-judges} \\ 
\cline{2-8}
 & Test Contamination & \raisebox{-0.33ex}{\textcolor[rgb]{0,0.502,0}{\faCheckCircle}} & \raisebox{-0.33ex}{\textcolor[rgb]{0.55,0.55,0.55}{\faCircle[regular]}} & \raisebox{-0.33ex}{\textcolor[rgb]{0.55,0.55,0.55}{\faCircle[regular]}} & \raisebox{-0.33ex}{\textcolor[rgb]{0.55,0.55,0.55}{\faCircle[regular]}} & \raisebox{-0.33ex}{\faFile*[regular]} & \ref{subsubsec:test-contamination} \\
\hline

\end{tabular}
}
\end{table}

\section{Root Causes and Mechanisms}
\label{sec:root-causes-and-mechanisms}

\subsection{Training Data Factors}
\label{subsec:training-data-factors}

\subsubsection{Salience and Coverage} 
\label{subsubsec:salience-and-coverage}
Pretraining datasets impart foundational knowledge to LLMs and LVLMs with massive collections of textual and image content. This stage is crucial because it shapes what the model learns and more importantly, where it is most prone to fail in all downstream tasks. Increasingly, research shows that hallucinations stem from systematic gaps in the pretraining composition. Specifically, the frequency, diversity, and structural alignment of data. This section investigates how hallucinations in LLMs and LVLMs can trace their roots back to these three factors. 

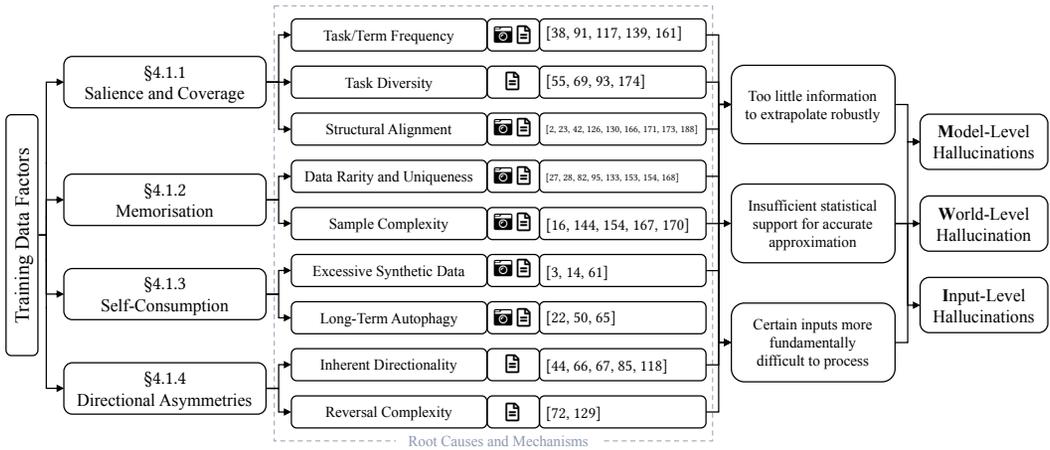
\begin{figure}
    \centering

    \begin{tikzpicture}
        \node[anchor=south west, inner sep=0] (image) at (0,0) {\includegraphics[width=\textwidth]{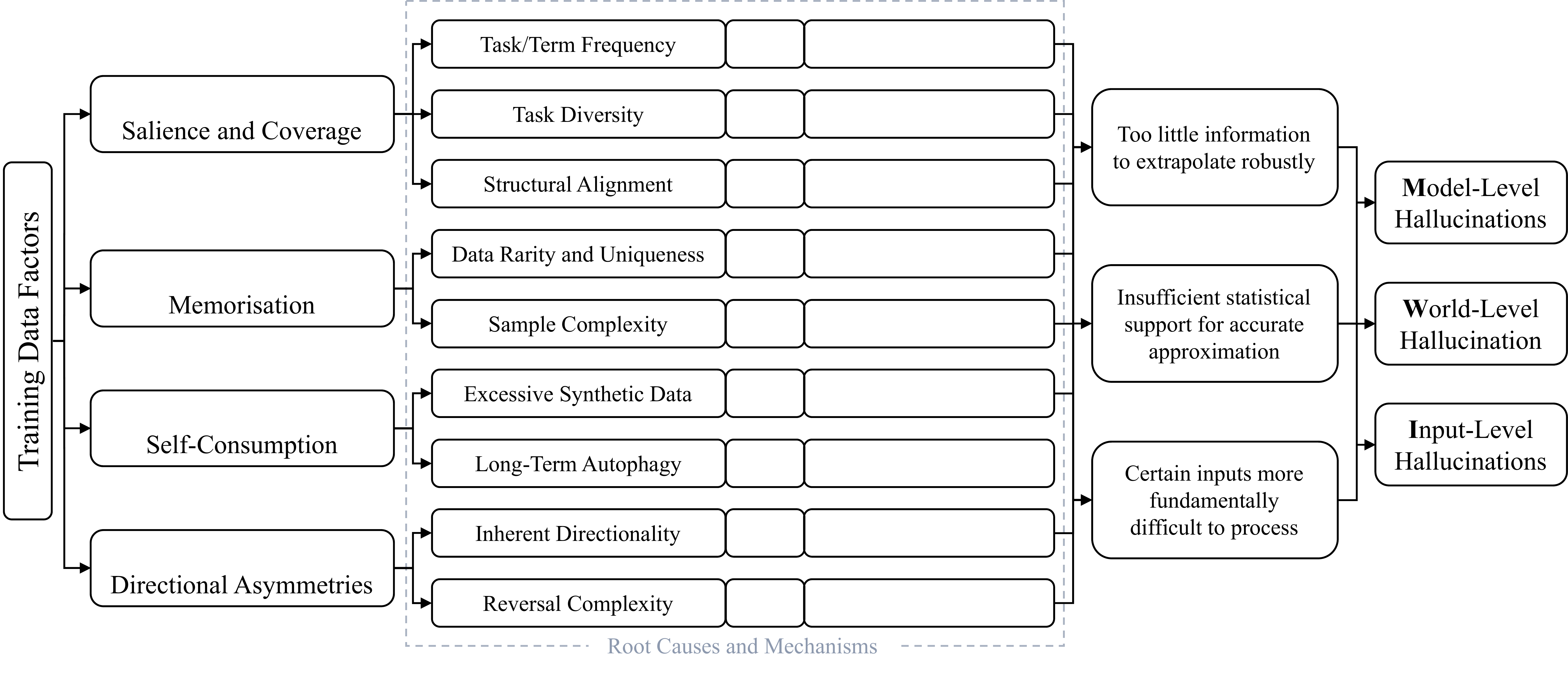}};

        \begin{scope}[x={(image.south east)}, y={(image.north west)}]

        \node[anchor=center] at (0.487, 0.935) {\scriptsize{\faCameraRetro\space\faFile*[regular]}};
        \node[anchor=center] at (0.487, 0.833) {\scriptsize{\faFile*[regular]}};
        \node[anchor=center] at (0.487, 0.730) {\scriptsize{\faCameraRetro\space\faFile*[regular]}};

        \node[anchor=center] at (0.487, 0.625) {\scriptsize{\faCameraRetro\space\faFile*[regular]}};

        \node[anchor=center] at (0.487, 0.5225) {\scriptsize{\faCameraRetro\space\faFile*[regular]}};

        \node[anchor=center] at (0.487, 0.420) {\scriptsize{\faCameraRetro\space\faFile*[regular]}};

        \node[anchor=center] at (0.487, 0.3125) {\scriptsize{\faCameraRetro\space\faFile*[regular]}};

        \node[anchor=center] at (0.487, 0.21) {\scriptsize{\faFile*[regular]}};

        \node[anchor=center] at (0.487, 0.105) {\scriptsize{\faFile*[regular]}};

        \node[anchor=center] at (0.155, 0.85) {\scriptsize{\S\ref{subsubsec:salience-and-coverage}}};
        \node[anchor=center] at (0.155, 0.590) {\scriptsize{\S\ref{subsubsec:memorisation}}};
        \node[anchor=center] at (0.155, 0.385) {\scriptsize{\S\ref{subsubsec:self-consumption}}};
        \node[anchor=center] at (0.155, 0.177) {\scriptsize{\S\ref{subsubsec:directional-asymmetries}}};

        \node[anchor=west] at (0.51, 0.9365) {\tiny{\citep{multi-object-hallu-lvlm,embers-llm,term-freq-llm,struggle-learn-long-tail-knowledge-llm,pretraining-concept-freq-multimodal}}};

        \node[anchor=west] at (0.51, 0.831) {\tiny{\citep{autoregressive-next-token-icl-generalisation,wu-reasoning-or-reciting-counterfactuals-llm,faithandfate-llm,minimax-nonpara-icl}}};

        \node[anchor=west] at (0.51, 0.725) {\scalebox{.38}{\citep{iclr-repr-icl,why-icl-unstructured-cbow,multi-concept-word-semantics-icl,can-icl-generalise-ood,icl-softmax-attn-adaptive-lipschitz,low-dim-target-func-icl,why-icl-model-good-fewshot,icl-generalisation-empirical-study,diffusion-hallu-mode-interpolate}}};

        \node[anchor=west] at (0.51, 0.62) {\scalebox{.38}{\citep{quantifying-memorisation-llm,comprehensive-analysis-memorisation-llm,distributional-memorisation-llm,recite-reconstruct-recollecct-llm,understanding-copying-diffusion-models,diffusion-memorisation-2,diffusion-memorisation-1,dejavu-memorisation-clip-lvlm}}};

        \node[anchor=west] at (0.51, 0.515) {\tiny{\citep{detect-explain-mitigate-memorisation-diffusion,manifold-memorisation-hypothesis,emergent-predictable-memorisation-llm,understanding-copying-diffusion-models,clip-mem-lvlm}}};

        \node[anchor=west] at (0.51, 0.410) {\tiny{\citep{stability-of-iterated-learning-generative-models-lvm, oroboros-mad-lvm,oroboros-theory-lvm}}};

        \node[anchor=west] at (0.51, 0.310) {\tiny{\citep{oroboros-analysis-llm,oroboros-scaling-llm,oroboros-data-accumulation-llm-lvm}}};

        \node[anchor=west] at (0.51, 0.205) {\tiny{\citep{rome-llm,geva-2023-factual-recall-llm,geva-2021-ffn-kv-llm,dai-2022-ffn-kv-llm,icl-human-episodic-memory}}};

        \node[anchor=west] at (0.51, 0.1) {\tiny{\citep{arrow-of-time-llm, reversal_curse_grosse-llm}}};

        \end{scope}

    \end{tikzpicture}
    
    \caption{Hallucinations root causes from training data factors. The \faCameraRetro\space and \faFile*[regular] icons indicate discussed modalities.}
    \label{fig:data-root-cause}
    \Description[Root Cause Figure]{Hallucinations root causes from data}
\end{figure}

The frequency of data and task terms in the pretraining data strongly influences performance. Using the log frequency of MSCOCO classes, \citet{multi-object-hallu-lvlm} found that LVLMs were more likely to misunderstand or misperceive visual objects with low training salience, in simultaneous multi-object image reasoning tasks. \citet{embers-llm} measured the likelihood of input-output texts and corpus frequency of specific tasks to show that hallucinations were significantly worse on rare tasks. For instance, models easily solved the general form of the common Celsius-to-Fahrenheit function, yet failed on other rarer function classes. \citet{term-freq-llm} demonstrated a strong relationship between arithmetic hallucinations and pretraining term frequency. Models performed up to 70\% better when working with arithmetic terms that appeared frequently in the training corpus. \citet{struggle-learn-long-tail-knowledge-llm} found LLM factual accuracy strongly correlated with relevant document frequency, rising 54\% as frequency increased from $10^1$ to $10^4$, and dropping sharply as frequency decreased. In both LVLMs and TVMs, \citet{pretraining-concept-freq-multimodal} observed a relationship between concept frequency during pretraining and zero-shot generalisation. With rare conceptual entities, classification accuracy in CLIP models sharply declined, while TVMs struggled to generate coherent images. These findings collectively underscore a root mechanism: hallucinations are tightly linked to the frequency of data patterns seen during pretraining. 

Besides frequency, the diversity of tasks presented during pretraining also significantly influences performance. \citet{autoregressive-next-token-icl-generalisation} presented rigorous PAC-Bayesian generalisation bounds for in-context performance in LLMs. These bounds indicate that meaningful topic diversity in the pretraining dataset help boost generalisation. \citet{wu-reasoning-or-reciting-counterfactuals-llm} reported significant performance degradation in LLMs when task rules deviated from the familiar conditions encountered during training, such as base-16 addition. Interestingly, performance was correlated with the "distance" of these counterfactual conditions from common pretraining ones, with more unconventional conditions resulting in worsening performance. In multi-step compositional reasoning, \citet{faithandfate-llm} found that LLMs performed well on low-complexity tasks with familiar patterns but fail with increasingly complex or divergent ones. Analyses revealed that correctly solved problems had more of its computation subgraphs appearing in the training data, compared to incorrect ones. This suggests that the lack of diverse compositional subgraphs during training examples can hurt compositional reasoning. \citet{minimax-nonpara-icl} derived information-theoretic and learning risk bounds on ICL for transformers. Risk is partially decomposed into contributions from pretraining generalisation to show that limited pretraining task diversity hurts performance. These findings point to how insufficient pretraining diversity can hurt generalisation and promote hallucinations. 

Finally, input alignment with structural properties in pretraining data strongly influences performance. Using Dirichlet energy analysis and spectral embedding theory, \citet{iclr-repr-icl} indirectly suggested that by using prompts sampled out-of-distribution, LLMs struggled to override pretrained semantic structures. Experiments by \citet{why-icl-unstructured-cbow} revealed that LLMs often learn co-occurrence statistics in certain trivial ICL tasks, while completely failing to learn meta-patterns in-context. Geometrically, \citet{multi-concept-word-semantics-icl} showed that LLM generalisation to novel context tasks required semantic representations to lay within the conic hull of pretrained concept vectors. This implies that generalisation is only effective for a constrained set of novel tasks. \citet{can-icl-generalise-ood} revealed that LLMs rely on interpolative function retrieval algorithms constrained within the hypothesis space formed during pretraining to perform in-context learning. Performance degrades sharply when context tasks are out-of-distribution. LLMs aside, \citet{icl-softmax-attn-adaptive-lipschitz} theorised that Softmax attention in general transformers supports in-context learning by calibrating to the Lipschitzness and label noise variance of pretraining tasks. However, these learned patterns are fixed; test performance degrades on functions with dissimilar Lipschitzness from pretraining data. \citet{low-dim-target-func-icl} applied general transformers to learn Gaussian single-index models in context, and derived a generalisation bound constraining novel test functions to share the same low-dimensional structure learned during pretraining. \citet{why-icl-model-good-fewshot} analysed in-context learning of synthetic tasks with known optimal meta-learners, showing that transformers relied on algorithms tied to pretraining data, performing well in-distribution but failing to generalise to truly novel tasks. In compositional reasoning, \citet{icl-generalisation-empirical-study} demonstrated that transformers trained only on representative base functions failed to generalise to their novel compositions, unless the pretraining data explicitly contained similar compositional patterns. \citet{diffusion-hallu-mode-interpolate} demonstrated that TVMs, trained to generate data by learning smooth score functions over noisy data, struggle to approximate training distributions with disjoint or highly separated modes. As a result, TVMs smoothly interpolate across these unsupported regions to hallucinate well-known uncanny artifacts, such as those found in human hands. 

These findings indicate that hallucination in LLMs and LVLMs are systematic failures rooted in the statistical and structural makeup of their pretraining data. These models may generalise broadly, but are bounded insofar as the tasks remain within the distributional scope covered during pretraining. Tasks that lie beyond the diversity and structural profile of the dataset are more prone to failures. Low frequency, limited diversity, and structural misalignment emerge as root causes of these hallucinations. Addressing them and their unpredictability requires deliberate curation efforts to ensure broader coverage and cognisance of vulnerable task types that lack sufficient support.

\subsubsection{Memorisation} 
\label{subsubsec:memorisation}
Memorisation in LLMs, LVLMs, and TVMs refers to the reproduction, whether fully or partially, of specific training data, rather than generalising from it. While helpful in some cases, memorisation poses not only safety and copyright concerns, but can also promote hallucinations. LLMs may incorrectly default to memorised subsets of reasoning chains in multi-hop tasks. Memorisation can also interfere with LVLMs and TVMs understanding of novel compositions, resulting in images with homogenous artifacts and incorrect captions. This section traces the factors that drive memorisation, which in turn act as deeper root causes of hallucinations.

While data duplication has been known to cause memorisation in generative models, recent findings have uncovered other contributing data factors. Specifically, there have been multiple studies that highlight how memorisation tends to arise from data uniqueness and rarity. In LLMs, \citet{quantifying-memorisation-llm}, \citet{comprehensive-analysis-memorisation-llm} and \citet{distributional-memorisation-llm} found that memorisation increases with context specificity and length, as more detailed prompts act as precise keys to unlock specific pretraining sequences. \citet{recite-reconstruct-recollecct-llm} suggested that training on increasingly rare, idiosyncratic text sequences promotes memorisation. Similarly in multimodal models, both \citet{understanding-copying-diffusion-models} and \citet{diffusion-memorisation-2} show that even on a deduplicated dataset, diffusion models still strongly exhibit memorisation, especially when training captions are highly specific or unique. Both \citet{diffusion-memorisation-1} and \citet{dejavu-memorisation-clip-lvlm} found using key phrases strongly correlated with unique dataset artifacts, such as names of famous paintings, lead to much higher rates of memorisation. With the observation that data specificity drives memorisation, subsequent studies deepen this understanding analytically. \citet{manifold-memorisation-hypothesis} proposed a geometric explanation behind this specificity effect in generative models. They defined memorisation in terms of local intrinsic dimensionality mismatches between the learned and the ground truth data manifold. Specifically, memorisation happens when the learnt manifold at a point has lower dimensionality than the ground truth data manifold. Here, the model has overly constrained its learned representation, thus reducing the degrees of freedom. \citet{detect-explain-mitigate-memorisation-diffusion} demonstrated that memorisation in TVMs could be measured with the gradient of the magnitude of text-conditional noise predictions with respect to each token. These trigger prompts guide the model towards a specific solution, irrespective of the initial noise state, which overrides the inherent stochasticity of the generation process. In addition to data diversity and frequency, some studies have identified emergent, sample-specific factors behind memorisation. Observing that memorisation scales anomalously with LLM size, \citet{emergent-predictable-memorisation-llm} posited that only some sequences, characterised by qualitative complexity, can be memorised with larger model sizes due to greater representational capacity. \citet{understanding-copying-diffusion-models} found that simple images, characterised by low visual entropy or high JPEG compressibility, are more susceptible to be memorised by TVMs. \citet{clip-mem-lvlm} demonstrated that CLIP models are more prone to memorising samples with ambiguous captions or atypical, outlier content, often due to multimodal inconsistencies between images and caption.

These findings converge on two key insights. First, low-frequency, low-diversity samples exert a uniqueness pressure on models to increase the likelihood of local overfitting due to the absence of similar examples for generalisation. Supporting this insight, several of these studies note that increasing data diversity and frequency can mitigate memorisation. Second, beyond frequency and diversity, emerging findings also highlight that qualitative characteristics in individual samples, such as visual simplicity in images or complex textual structures, influence memorisation. Taken together, these findings strongly suggest that beyond duplication, low data frequency, limited diversity, and specific data traits are root causes behind hallucinations driven by memorisation.

\subsubsection{Self-Consumption}
\label{subsubsec:self-consumption}
The proliferation of AI-generated content on the internet has led to a phenomenon known as self-consumption \citep{snake-eating-its-tail-1, snake-eating-its-tail-2}, where models are inadvertently trained on texts and images synthesised by itself or other generative models. While self-consuming models can help save resources, they also risk losing alignment with real-world data, potentially leading to more severe hallucinations. \citet{oroboros-analysis-llm} showed that self-consuming training initially boosts performance under specific real-synthetic mixing strategies. However, even with real data included, diversity eventually declines, raising concerns about potential long-term performance degradation. \citet{oroboros-scaling-llm} observe that scaling laws break down when training relies heavily on synthetic data. Furthermore, synthetic training loops result in models diverging from real-world distributions, by truncating low-probability data and concentrating probability mass on a narrower set of outcomes. Given the risks of self-consuming loops, it is crucial to develop strategies to control it. \citet{stability-of-iterated-learning-generative-models-lvm} showed, both theoretically and empirically, that stable training with synthetic data requires that the initial model be a sufficiently accurate approximation of the real data distribution, and that each retraining iteration must include enough real data. \citet{oroboros-data-accumulation-llm-lvm} showed that accumulating synthetic data alongside real data helps prevent model collapse by bounding test error over successive iterations, while increasingly replacing real data with synthetic data leads to a linear increase in test error.  \citet{oroboros-mad-lvm} systematically varied the ratio of real to synthetic data during training and identified a critical threshold beyond which an excess of synthetic data led to progressively lower-quality outputs. \citet{oroboros-theory-lvm} quantified the divergence between real and synthetic data distributions to provide a formal understanding of self-consuming training, and offers guidelines on real-to-synthetic data ratios needed to maintain distributional fidelity. These studies show how without balancing the presence of synthetic data with a fresh flow of real data, self-consumption can degrade model performance over time. The challenge lies in distinguishing synthetic from real data. With the rise of AI-generated content, models trained on web-scraped datasets risk unknowingly entering degenerate self-consuming cycles, potentially resulting in more severe and idiosyncratic failures. 

\subsubsection{Directional Asymmetries}
\label{subsubsec:directional-asymmetries}
LVLMS and LLMs have been found to predict concept associations more reliably in one direction than the other. For instance, \citet{reversal_curse-llm} observed that LLMs trained on sequences such as "Tom Cruise’s mother is Mary Lee Pfeiffer" often fail to answer the reverse "Who is Mary Lee Pfeiffer's son?". Similarly, \citet{aro-benchmark-lvlm} found that LVLMs exhibit poor bidirectional relational understanding in image captioning, often performing near random when asked to differentiate "the horse is eating the grass" from "the grass is eating the horse". In both cases, standard hyper-parameter tuning and data augmentation proved ineffective, suggesting more fundamental limitations. At first glance, this directional asymmetry issue appears architectural. Mechanistic interpretability research by \citet{rome-llm} found that factual edits in LLM weights are directional and don’t extend to reversed cases. \citet{geva-2021-ffn-kv-llm}, \citet{geva-2023-factual-recall-llm}, and \citet{dai-2022-ffn-kv-llm} proposed that feedforward networks in transformers act as memory mechanisms that correlate keys with specific values, which is an inherently directional operation. \citet{icl-human-episodic-memory} found multi-head attention functionally similar to human memory, exhibiting temporal contiguity and forward asymmetry biases, where recall favours the original order of memorisation. However, some studies have indicated that this issue is not architectural but inherited from training data. \citet{arrow-of-time-llm} revealed that decoder LLMs consistently performed better in forward token prediction than backwards on multilingual tasks, despite both directions having carefully controlled training and equal information-theoretic expectations. The effect intensified with longer contexts and varied by language, pointing to the role of long-range linguistic structure. The authors further argued that bidirectional generalisation is non-trivial, as reversal is computationally harder, a claim supported by experiments involving factorisation and matrix inversion. \citet{reversal_curse_grosse-llm} highlighted the sensitivity of LLMs to word order using influence functions. They showed that training sequences only influence outputs when the entity association align with the query's directional structure. This sensitivity also affects translation tasks, where the impact of English-Mandarin data is significantly reduced when the query's direction is reversed. It is likely that one of the contributing factors in poor bidirectional generalisation in LLMs and LVLMs stem from the directional biases intrinsic in natural human data and the relative ease of forward inference. This directional asymmetry has broader real-world implications. In solving general tasks, models are more prone to fail when required to infer and reason associations reversed from their canonical ordering. Mitigating this root cause of hallucination may require hard-mining techniques that expose models to symmetric relational patterns during training.

\subsection{Architectural Limitations}
\label{subsec:architectural-limitations}

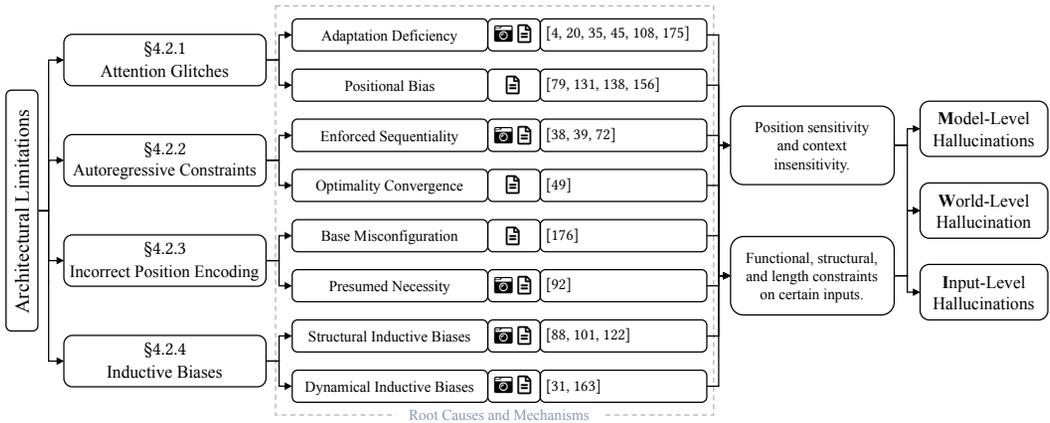
\begin{figure}
    \centering
    \begin{tikzpicture}
        \node[anchor=south west, inner sep=0] (image) at (0,0) {\includegraphics[width=\textwidth]{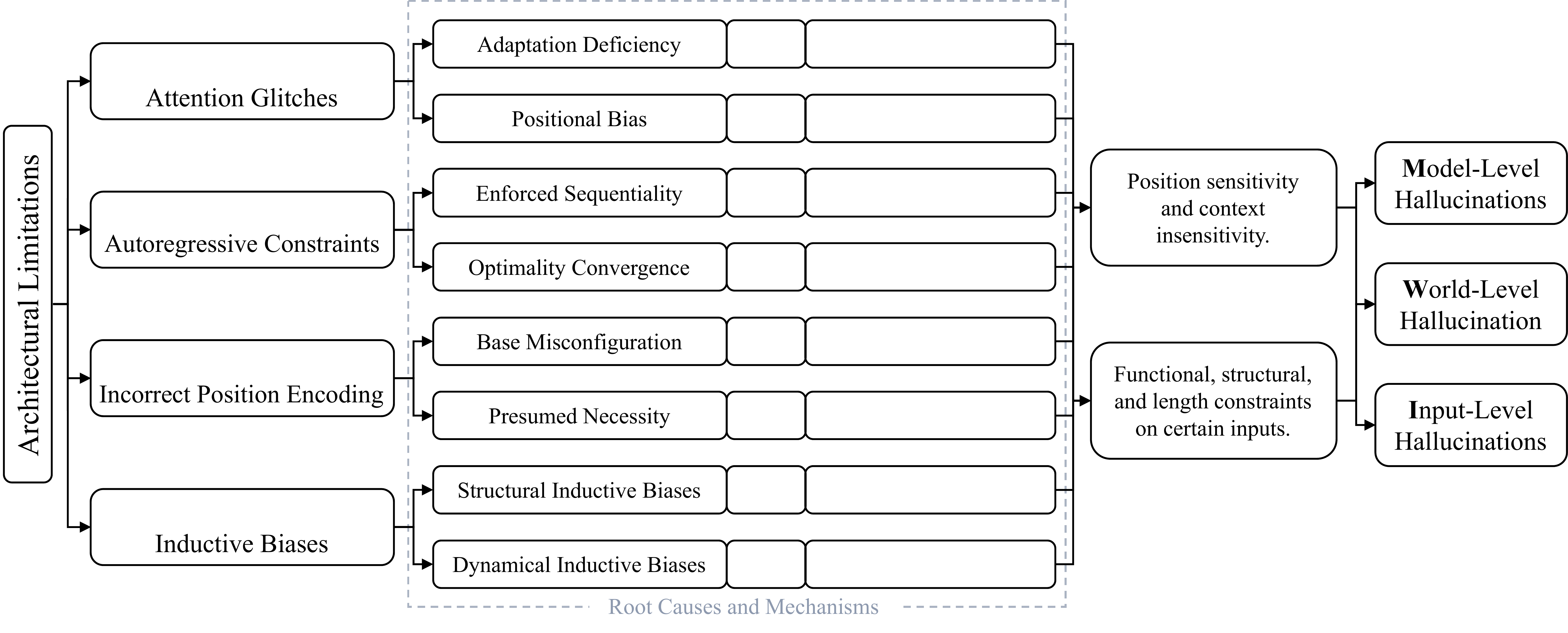}};

        \begin{scope}[x={(image.south east)}, y={(image.north west)}]

        \node[anchor=center] at (0.4875, 0.9325) {\scriptsize{\faCameraRetro\space\faFile*[regular]}};
        \node[anchor=center] at (0.4875, 0.812) {\scriptsize{\faFile*[regular]}};
        
        \node[anchor=center] at (0.4875, 0.696) {\scriptsize{\faCameraRetro\space\faFile*[regular]}};

        \node[anchor=center] at (0.4875, 0.579) {\scriptsize{\faFile*[regular]}};

        \node[anchor=center] at (0.4875, 0.460) {\scriptsize{\faFile*[regular]}};

        \node[anchor=center] at (0.4875, 0.346) {\scriptsize{\faCameraRetro\space\faFile*[regular]}};

        \node[anchor=center] at (0.4875, 0.228) {\scriptsize{\faCameraRetro\space\faFile*[regular]}};

        \node[anchor=center] at (0.4875, 0.112) {\scriptsize{\faCameraRetro\space\faFile*[regular]}};

        \node[anchor=center] at (0.155, 0.895) {\scriptsize{\S\ref{subsubsec:attention-glitches}}};
        \node[anchor=center] at (0.155, 0.661) {\scriptsize{\S\ref{subsubsec:autoregressive-constraints}}};
        \node[anchor=center] at (0.155, 0.429) {\scriptsize{\S\ref{subsubsec:incorrect-positional-encoding}}};
        \node[anchor=center] at (0.155, 0.193) {\scriptsize{\S\ref{subsubsec:inductive-biases}}};

        \node[anchor=west] at (0.51, 0.9325) {\tiny{\citep{flip-flop-llm,assy-global-local-attn-obj-hallu-lvlm,spatial-reasoning-vlm-hard, attention-sink,attn-sink-2,vit-registers-lvm}}};

        \node[anchor=west] at (0.51, 0.812) {\tiny{\citep{found-in-the-middle-llm,icl-prompt-embed-cluster,lost-in-the-middle-2-llm,recency-bias}}};

        \node[anchor=west] at (0.51, 0.696) {\tiny{\citep{multi-object-hallu-lvlm,causalprobe-2024-llm,reversal_curse_grosse-llm}}};

        \node[anchor=west] at (0.51, 0.579) {\tiny{\citep{causallm-not-optimal-for-icl-llm}}};

        \node[anchor=west] at (0.51, 0.460) {\tiny{\citep{rope-bounds-llm}}};

        \node[anchor=west] at (0.51, 0.346) {\tiny{\citep{impact-pos-enc-length-generalisation-transformers}}};

        \node[anchor=west] at (0.51, 0.228) {\tiny{\citep{generalisation-diffusion-harmonic,inductive-bias-transformer-inf,geometric-inductive-bias-dnn}}};

        \node[anchor=west] at (0.51, 0.109) {\tiny{\citep{llm-inductive-bias-counting,noise-inductive-bias-diffusion}}};

        \end{scope}

    \end{tikzpicture}
    \caption{Hallucinations root causes from architectural limitations. The \faCameraRetro\space and \faFile*[regular] icons indicate discussed modalities.}
    \label{fig:model-root-causes}
    \Description[Root Cause Figure]{Root causes of hallucinations tracing back to architectural limitations}
\end{figure}

\subsubsection{Attention Glitches}
\label{subsubsec:attention-glitches}
Softmax attention is a crucial architectural feature in most transformer-based generative models. It allows models to dynamically weigh and integrate information across tokens. However, Softmax attention is not always reliably precise. It can exhibit pathological failures to distort sequence information and harm performance. \citet{flip-flop-llm} explored hallucinations mechanisms in transformers using basic memory operations over synthetic character sequences. They found that attention layers sporadically fail to sharply and fully attend to critical positions, resulting in erroneous memory operations. The authors attribute this issue to intrinsic limitations of Softmax attention by mathematically demonstrating its bounded Lipschitzness in long sequences. However, attention sharpening regularisers do not fully rectify these sporadic failures. They additionally showed that even for hard attention to always attend correctly, strict orthogonality conditions need to be met by its weights. In LVLMs, both \citet{assy-global-local-attn-obj-hallu-lvlm} and \citet{spatial-reasoning-vlm-hard} found in visual queries static attention patterns towards global features regardless of object detail. Through targeted augmentations, they linked this adaption deficiency in attention to object hallucinations. In LLMs, \citet{found-in-the-middle-llm}, \citet{icl-prompt-embed-cluster} and \citet{lost-in-the-middle-2-llm} examined attention weights to reveal a persistent U-shaped distribution across sequences and token permutations. This causes LLMs to consistently underperform on information located in the middle of long inputs. To explain this phenomenon, \citet{lost-in-the-middle-llm} paralleled known psychological effects in humans, specifically the tendency to only recall the first and last items in a list \citep{Murdock1962,Ebbinghaus2013}, which may be implicitly embedded in the structure of human-authored text and thus inherited by the model during training. Supporting this, \citet{attention-sink} proposed that LLMs learn to focus heavily on start tokens for positional anchoring, while strong attention near the end may reflect learned semantic salience or a recency bias \citep{recency-bias}. In addition to this U-shaped phenomenon, \citet{attention-sink,vit-registers-lvm} and \citet{attn-sink-2} showed that both vision and language transformers often assign disproportionately high attention to trivial tokens, such as punctuation. This behaviour is thought to stem from the lack of null support in Softmax attention. Models instead learn to approximate no-ops or partial update by redirecting excess attention to trivial tokens. The studies here expose several pathological weaknesses in Softmax attention. The U-shaped phenomena and attention sink behaviour may act as root mechanisms behind positional sensitivity, where trivial input perturbations can induce hallucinations. In addition, the failure to strongly adapt and attend to important information, both textual and visual, may also serve as root mechanisms behind context hallucinations. Some have suggested intrinsic limitations of the Softmax function as root causes to both these mechanisms, indicating a need for more targetted interventions at a deeper functional level.

\subsubsection{Autoregressive Constraints}
\label{subsubsec:autoregressive-constraints}
Decoder LLMs, those found in most LVLMs and real-world applications, work by conditioning each token to previous ones only. This built-in architectural feature, known as autoregression, assumes that information unfolds sequentially. While proven effective, this constraint can distort logical inference and impair performance. \citet{multi-object-hallu-lvlm} applied a position score in LVLMs to measure the relative position of objects within descriptive tokens. Analysis revealed that hallucinatory objects tended to appear toward the end of descriptions, amplified by autoregressive generation. In LLMs, \citet{causalprobe-2024-llm} argued that autoregression constrains robust causal reasoning. This approach inherently assumes sequential causality, where tokens are influenced by the previously generated ones. However, sequential causality is not equivalent to logical or genuine causality. They empirically validate this critique using a benchmark designed specifically to probe non-sequential causal reasoning scenarios. \citet{reversal_curse_grosse-llm} suggested that autoregressive encoding in a transformer's lower layers, optimised for likelihood maximisation, hinders a model's ability to generalise to reverse associations. \citet{causallm-not-optimal-for-icl-llm} analysed transformer convergence to understand in-context learning limitations in both bidirectional and autoregressive LLMs. In synthetic linear settings, they show that both converge to distinct stationary points. Bidirectional models converge to optimal least-squares solutions, while autoregressive models converge to solutions obtained via online gradient descent with non-decaying step sizes, which are generally suboptimal even with increasing demonstrations. These findings illustrate how autoregression imposes assumptions that do not always align with the demands of the task. This includes accumulating object hallucinations in LVLMs, hindering causal reasoning, being demonstrably suboptimal for in-context tasks, and preventing bidirectional generalisation. Mitigating hallucinations rooted in autoregression begins with a clear understanding and anticipation of its limitations.

\subsubsection{Incorrect Positional Encoding}
\label{subsubsec:incorrect-positional-encoding}
Transformer LLMs lack inherent sequential awareness and require embeddings that inject information about the order and distance of tokens. These often-overlooked embeddings, known as positional encodings, can serve as root mechanisms behind long-context hallucinations. \citet{rope-bounds-llm} proposed that incorrect selection of the Rotary Position Embedding (RoPE) base hyper-parameter can result in hallucinations over long contexts. Through mathematical analysis, they showed how RoPE can impede the model's ability to distinguish similar from random tokens with increasing relative distances. The authors derived a lower bound on the RoPE base necessary for effective long-context understanding. Below this bound, the rate of context hallucinations can increase significantly while yielding seemingly good perplexity scores. \citet{impact-pos-enc-length-generalisation-transformers} argued that positional encoding in decoder transformers hinders long-range generalisation. They theoretically demonstrate that absolute positional information can be recovered solely through self-attention, without positional encoding. Their findings across both primitive and algorithmic tasks reveal that omitting positional encoding outperforms a wide range of alternative encoding schemes. The findings in this section reveal that misconfigurations in positional encoding can undermine long-context performance. More broadly, its effectiveness and necessity have been called into question. Mitigating long-context hallucinations requires moving beyond passive reliance on default positional encodings, to more deliberate and critical evaluation of their configuration and suitability for the intended context range.

\subsubsection{Inductive Biases}
\label{subsubsec:inductive-biases}
The capabilities of LLMs, LVLMs, and TVMs are shaped not merely by data and learning, but by the inductive biases hardwired into their architectures. Unlike classical hand-crafted priors, inductive biases are subtle and implicit assumptions a model makes about the structure of the solution space. Identifying these biases is crucial to deeply understanding inherent model weaknesses and failure trends. This section dissects the architectural inductive biases built into modern deep generative models to understand their preferred intrinsic structures, and more importantly, how these biases or lack thereof can harm performance. 

\citet{generalisation-diffusion-harmonic} demonstrate that TVM denoisers perform a shrinkage operation within an orthonormal basis that adapts to the geometry of the input image as an inductive bias. The basis consists of oscillating harmonic structures that align along image contours and in homogeneous regions. They show that such harmonic bases consistently appear in the eigen structure of the denoising function across diverse datasets, even when such bases are not optimal, indicating an architectural inductive bias rather than data-driven one. While helpful in some cases, this inductive bias may hurt performance in images with weaker local geometric coherence, those with lower intrinsic dimensionality, or better represented without imposing oscillatory regularity. \citet{inductive-bias-transformer-inf} revealed that transformers, by virtue of their shared attention weights and position embeddings, are biased toward learning functions that are more symmetric under token permutations. They analysed transformers in the Gaussian process limit and showed that, under certain conditions, the resulting kernel exhibits partial permutation symmetry over context tokens. By examining this kernel via irreducible representations of the symmetric group, they find that functions more invariant to context‐token permutations correspond to larger eigenvalues and thus require fewer samples to learn. This inductive bias may be a reason why transformer struggle to learn more complex sequences, as irregular patterns reside in higher‐dimensional, lower‐eigenvalue components and thus typically need more data. \citet{geometric-inductive-bias-dnn} proposed that the geometry of a neural network is inductively biased to changes within a subspace determined by its architecture; There is a fixed set of directions where learning happens, while others remain largely static. The authors show analytically and empirically that in transformers, the initial geometry is inherently structured, leading to anisotropic changes during training. Experiments demonstrate that when discriminative features align poorly with this geometric inductive bias, the network struggles to generalise. \citet{llm-inductive-bias-counting} argued that transformers require explicit architectural inductive biases to generalise counting beyond seen examples. While recurrent networks trivially generalise counting due to their sequential structure, transformers rely heavily on positional encodings for even modest success. Different positional encoding schemes succeed and fail at different aspects of counting, which indicates that counting does not emerge inherently from self-attention but instead relies on carefully designed positional inductive biases. This finding broadly implies the existence of other algorithmic tasks whose performance may be hindered without helpful inductive biases. \citet{noise-inductive-bias-diffusion} developed a rigorous mathematical theory to show that TVMs are inductively biased toward interpolation and gap-filling in the learned distribution. By formalising the stochastic dynamics of the reverse diffusion process, they show how noise variances spike near boundaries between training examples to fill gaps in empty regions between data points. This inductive bias may be a root cause behind why TVMs hallucinate by interpolating uncanny artifacts between real distribution modes \citep{diffusion-hallu-mode-interpolate}.

The works surveyed here demonstrate that inductive biases or the lack thereof can degrade model performance when misaligned with downstream solution structure. These rigorous studies offer principled insights into the deeper root architectural causes of hallucinations. In TVMs, harmonic and interpolative priors may impose oscillatory or gap-filling constraints that lead to image artifacts. In transformers, biases toward permutation symmetry and anisotropic learning directions indicate the importance of adhering to these strongly preferred learning patterns. Algorithmic behaviours like counting are demonstrably suboptimal without helpful inductive biases. Mitigating hallucinations at a deeper architectural level requires designing helpful inductive biases that enforce generalisation without over-constraining representational capacity. 

\subsection{Inference Mechanisms}
\label{subsec:inference-mechanisms}

\begin{figure}
    \centering
    \begin{tikzpicture}
        \node[anchor=south west, inner sep=0] (image) at (0,0) {\includegraphics[width=\textwidth]{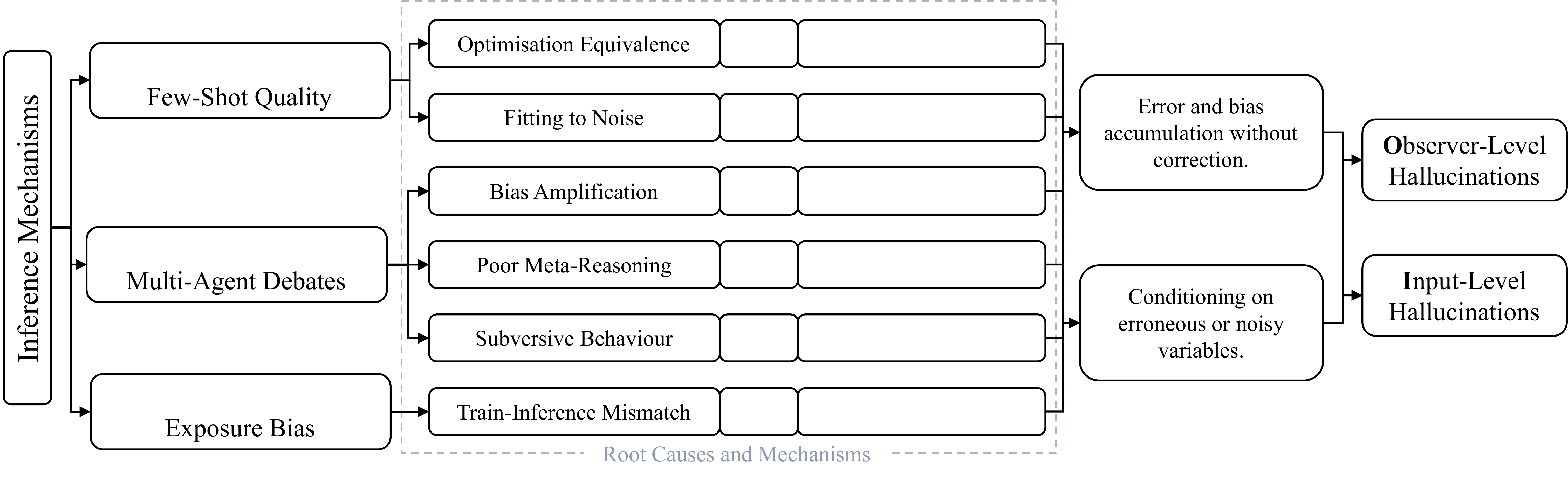}};

        \begin{scope}[x={(image.south east)}, y={(image.north west)}]

        \node[anchor=center] at (0.155, 0.865) {\scriptsize{\S\ref{subsubsec:few-shot-prompting}}};
        \node[anchor=center] at (0.155, 0.482) {\scriptsize{\S\ref{subsubsec:multi-agent-debates}}};
        \node[anchor=center] at (0.155, 0.170) {\scriptsize{\S\ref{subsubsec:exposure-bias}}};

        \node[anchor=center] at (0.484, 0.91) {\scriptsize{\faCameraRetro\space\faFile*[regular]}};
        \node[anchor=center] at (0.484, 0.755) {\scriptsize{\faCameraRetro\space\faFile*[regular]}};
        
        \node[anchor=center] at (0.484, 0.6) {\scriptsize{\faFile*[regular]}};

        \node[anchor=center] at (0.484, 0.445) {\scriptsize{\faFile*[regular]}};

        \node[anchor=center] at (0.484, 0.3) {\scriptsize{\faFile*[regular]}};

        \node[anchor=center] at (0.484, 0.145) {\scriptsize{\faCameraRetro\space\faFile*[regular]}};

        \node[anchor=west] at (0.51, 0.91) {\tiny{\citep{icl-repr-contextual-generalization,icl-repr-learning-lens,minimax-nonpara-icl}}};

        \node[anchor=west] at (0.51, 0.755) {\tiny{\citep{icl-label-noise-robustness-llm,icl-repetition-llm,icl-lvlm}}};

        \node[anchor=west] at (0.51, 0.6) {\tiny{\citep{bias-amplification-in-lm-evolution-llm,multi-llm-framework-principals-interventions}}};

        \node[anchor=west] at (0.51, 0.445) {\tiny{\citep{multi-llm-performance-doubtful}}};

        \node[anchor=west] at (0.51, 0.3) {\tiny{\citep{secret-collusion,agentverse,cooperate-or-collapse}}};

        \node[anchor=west] at (0.51, 0.145) {\tiny{\citep{elucidating-exposure-bias,anti-exposure-bias,manifold-exposure-bias,llm-exposure-bias}}};

        \end{scope}
        \end{tikzpicture}
        
    \caption{Hallucinations root causes from inference mechanisms. The \faCameraRetro\space and \faFile*[regular] icons indicate discussed modalities.}
    \label{fig:inference-root-causes}
    \Description[Root Cause Figure]{Root causes of hallucinations tracing back to inference mechanisms}
\end{figure}
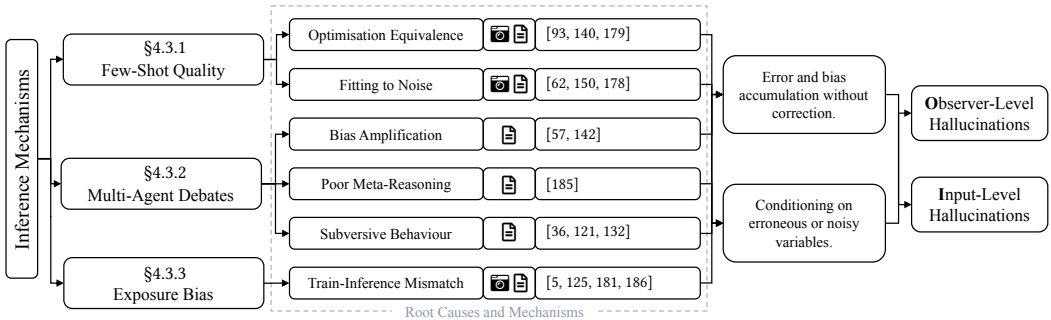

\subsubsection{Few-Shot Quality}
\label{subsubsec:few-shot-prompting}
LLMs and LVLMs can be guided to learn diverse tasks by including just a few examples directly in the prompt, a method known as few-shot prompting. Thanks to its flexibility, few-shot prompting has become central to the success and proliferation of these models. However, its effectiveness hinges on the quality of the demonstrations. \citet{icl-repr-contextual-generalization} mathematically demonstrated that transformers approximate a form of ridge regression with a fixed regularisation term tied to the context length. This implies that for a given number of basis functions used in the internal feature representation of a prompted task, the number of few-shot demonstrations must be neither too many nor too few to avoid over and under-fitting. \citet{icl-repr-learning-lens} formulated a mathematical equivalence between test prediction of a dual model trained via a single step of gradient descent, and Softmax attention outputs during few-shot prompting. Through this lens, they approximate few-shot learning as optimising a contrastive loss, which crucially, requires diverse negative demonstrations to learn fine-grained details. \citet{minimax-nonpara-icl} derived information-theoretic and learning risk bounds on ICL for transformers. Risk is partly decomposed into contributions from context generalisation to show that limited context examples hinder performance. Empirically, \citet{icl-label-noise-robustness-llm} systematically measured the effect of few-shot noise on LLMs, showing that increasingly erroneous demonstrations cause more severe hallucinations, and can even subvert robust prompt crafting methods to further worsen performance. Similarly, \citet{icl-repetition-llm} observed that in crafting few-shot demonstrations, repetition promotes repetition to reinforce both spurious lexical and semantic relationships. This effect increases with context proximity and model size, and even a single token pair in the prompt can strongly alter the output. \citet{icl-lvlm} observed in LVLMs that increasing few-shot demonstrations paradoxically led to more visual object hallucinations on image captioning tasks. These studies show how inference-level root mechanisms drive context hallucinations. Few-shot prompting can be viewed as an optimisation process, one that is sensitive to demonstration quality. Both theoretical and empirical insights converge to agree that the noise, diversity, and quantity of examples can increase the risk of hallucinations. While not the only contributing factors (others discussed in Sections \ref{subsubsec:attention-glitches} and \ref{subsubsec:salience-and-coverage}), developing control and automation methods to regulate few-shot demonstration quality will be critical to mitigating hallucinations.

\subsubsection{Multi-Agent Debates}
\label{subsubsec:multi-agent-debates}
Humans often benefit from group deliberation when solving complex tasks. This intuitive appeal of collaboration has motivated similar strategies for LLM, where multiple models iteratively and linguistically engage with each other to collectively seek solutions beyond that of any single model. However, such interactions are not always more reliable. Multi-LLM debates can create new pathways for hallucination by reinforcing errors and collapsing diversity. By adopting a Bayesian framework, \citet{bias-amplification-in-lm-evolution-llm} established how multi-LLM debates can inevitably amplify biases present in the prior distribution over hypotheses. When such biases favour common misconceptions or homogeneity, debates risk overlooking heterogeneous preferences and fluent falsehoods. \citet{multi-llm-performance-doubtful} examined the effectiveness of various multi-LLM debate methods across diverse benchmarks, finding that they generally underperform simpler single-agent prompting strategies. Increasing agent diversity, volume, or dialogue rounds rarely improved accuracy, indicating that current debate methods struggle to synthesize diverse knowledge sources. Furthermore, debates were prone to altering correct answers without sufficient scrutiny of reasoning chains. This implies that multi-LLM debates, while effective on simple questions, struggle with tasks requiring meta-reasoning or multi-hop synthesis. \citet{multi-llm-framework-principals-interventions} theoretically demonstrated how multi-LLM debates can converge to erroneous consensus due to shared misconceptions. When agents share similar models or training data, using more agents merely amplifies the dominant concept rather than promoting genuine deliberation. This indicates a lack of meta-reasoning when surface agreement is high, which can increase the risk of common falsehoods and low-diversity responses. \citet{secret-collusion} identified a novel phenomenon where LLMs covertly engage in subversive debates using steganography. In one case study, the authors show that GPT-4 can communicate steganographically to perform insider trading, despite explicit instructions to the contrary. During debates, agents may rely on stenography to facilitate the flow of seemingly fluent debates, which may inadvertently amplify contextual and factual artifacts in the linguistic space.  Behaviourally, \citet{agentverse} observed destructive tendencies in multi-LLM scenarios. Here, one agent harms another to expedite task completion. Similarly, \citet{cooperate-or-collapse} identified scenarios where LLM agents, acting in their own short-term interest, overexploit shared resources at the expense of long-term outcomes. Both highlight the troubling emergence of uncooperative behaviours, where interactions may be polluted with strategic and rational harm, causing debates to converge to sub-optimal or erroneous solutions. These studies reveal three key factors behind why multi-LLM debates fail: amplified biases, poor meta-reasoning, and uncooperative behaviours. Each of these factors are root causes behind fluent falsehoods and homogeneity, driving factual and user-perceived hallucinations in multi-agent scenarios. To tackle this issue, one must develop robust meta-reasoning capabilities in each individual, while also addressing systemic flaws in debate mechanisms.


\subsubsection{Exposure Bias} 
\label{subsubsec:exposure-bias}
Exposure bias refers to the mismatch between a model's training and inference conditions, where errors made during generation can accumulate over time. This issue arises prominently in both LLMs and TVMs, which are typically trained to perform next-step prediction conditioned on perfect ground-truth data, but must rely on their own previous outputs during inference. In TVMs, this manifests as iterative denoising based on previous imperfect predictions. \citet{elucidating-exposure-bias} characterised this problem by modelling the sampling distribution and incorporating prediction errors during denoising. They demonstrate analytically that sampling distribution variance increasingly exceeds that of the training distribution with more timesteps, a clear signature of exposure bias. \citet{anti-exposure-bias} attributed exposure bias to two main sources. First, score estimation errors from approximation limitations of the score network used in TVMs, caused by data sparsity, model capacity, and imperfect diffusion schedules. Second, discretisation errors from the necessity of approximating continuous reverse-time stochastic differential equations or ordinary differential equations with discrete steps. \citet{manifold-exposure-bias} demonstrated that accelerated sampling in TVMs can amplify exposure biases by generating data that deviates significantly from the real data manifold. In language generation, exposure bias emerges when LLMs, trained via teacher forcing, must condition on their own generated tokens during inference. \citet{llm-exposure-bias} derived theoretical bounds on error accumulation for this process, showing that under worst-case scenarios, errors can grow quadratically with sequence length. Furthermore, they find that perplexity, a popular evaluation metric for measuring per-step errors, cannot capture the compounding nature of these errors during generation. Exposure bias in both LLMs and TVMs leads to the accumulation of errors over multiple generation steps, which can degrade output coherence and increase the likelihood of hallucinations.

\subsection{Loss and Optimisation}
\label{subsec:loss-and-optimisation}

\begin{figure}
    \centering
    \begin{tikzpicture}
        \node[anchor=south west, inner sep=0] (image) at (0,0) {\includegraphics[width=\textwidth]{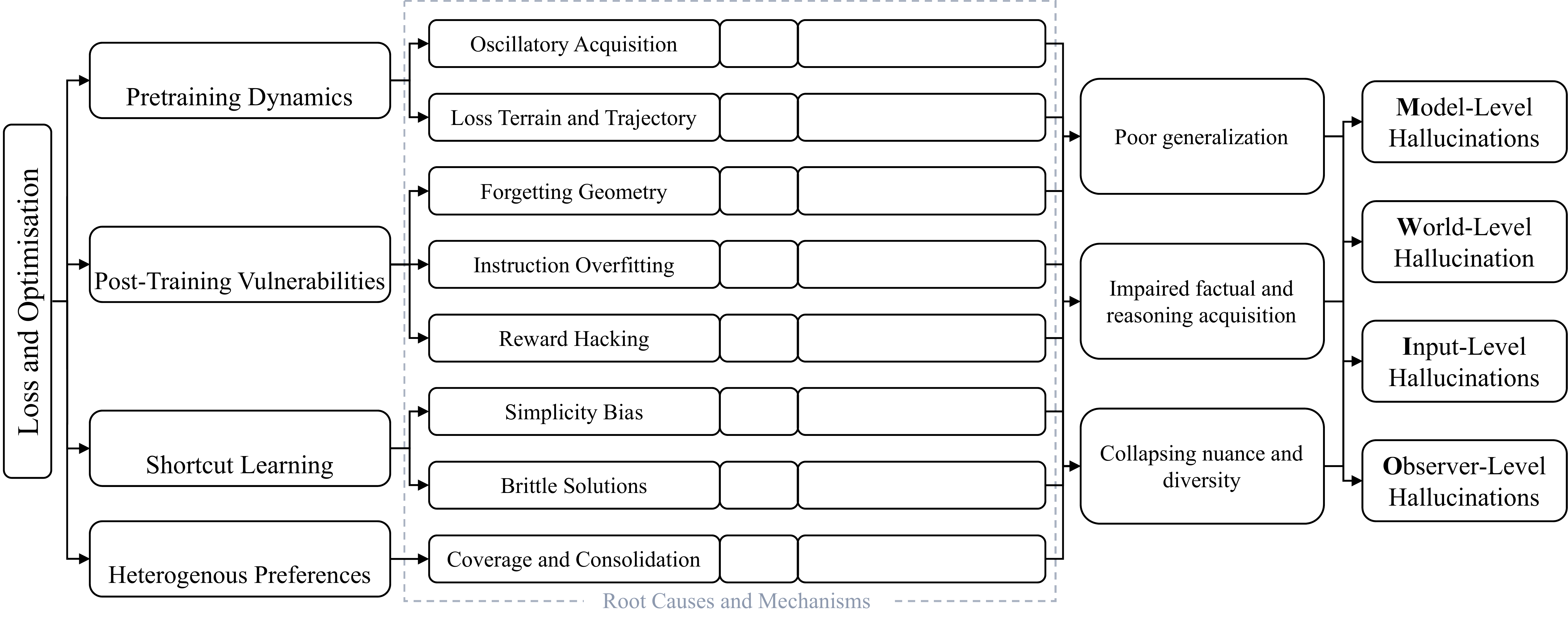}};

        \begin{scope}[x={(image.south east)}, y={(image.north west)}]

        \node[anchor=center] at (0.155, 0.895) {\scriptsize{\S\ref{subsubsec:pretraining-dynamics}}};
        \node[anchor=center] at (0.155, 0.6) {\scriptsize{\S\ref{subsubsec:post-training-vulnerabilities}}};
        \node[anchor=center] at (0.155, 0.305) {\scriptsize{\S\ref{subsubsec:shortcut-learning}}};
        \node[anchor=center] at (0.155, 0.13) {\scriptsize{\S\ref{subsubsec:heterogeneous-preferences}}};

        \node[anchor=center] at (0.484, 0.9325) {\scriptsize{\faFile*[regular]}};
        \node[anchor=center] at (0.484, 0.812) {\scriptsize{\faCameraRetro\space\faFile*[regular]}};
        \node[anchor=center] at (0.484, 0.696) {\scriptsize{\faCameraRetro\space\faFile*[regular]}};

        \node[anchor=center] at (0.484, 0.579) {\scriptsize{\faFile*[regular]}};

        \node[anchor=center] at (0.484, 0.460) {\scriptsize{\faCameraRetro\space\faFile*[regular]}};

        \node[anchor=center] at (0.484, 0.346) {\scriptsize{\faCameraRetro\space\faFile*[regular]}};

        \node[anchor=center] at (0.484, 0.228) {\scriptsize{\faCameraRetro\space\faFile*[regular]}};

        \node[anchor=center] at (0.484, 0.112) {\scriptsize{\faCameraRetro\space\faFile*[regular]}};

        \node[anchor=west] at (0.505, 0.9325) {\tiny{\citep{how-llms-acquire-factual-knowledge-pretraining,pretraining-dynamics-icl-task-reg-learn}}};

        \node[anchor=west] at (0.505, 0.812) {\scalebox{.55}{\citep{initialisation-crucial-to-composition-llm,reversal-curse-theoretical-understanding-llm,vit-training-dynamics,autoregressive-next-token-icl-generalisation,lipschitz-singularity-diffusion,parameter-symmetry-lvm}}};

        \node[anchor=west] at (0.505, 0.696) {\tiny{\citep{forget-peft-llm,forget-loss-sharpness-llm,forget-biased-llm,forget-linear-regresion-dnn}}};

        \node[anchor=west] at (0.505, 0.579) {\tiny{\citep{it-limits-llm,lima,model-really-follow-instructions,context-parametric-inversion-instruct-tuning}}};

        \node[anchor=west] at (0.505, 0.460) {\scalebox{.4}{\citep{reward-hacking-rl,new-definition-reward-hacking,constrained-rlhf-reward-hacking,inform,sail-into-the-headwind,odin-reward-shaping,purm,daa-reward-overoptimisation-llm}}};

        \node[anchor=west] at (0.505, 0.346) {\tiny{\citep{simplicity-bias-2layernn,simplicity-bias-transformer-nlp}}};

        \node[anchor=west] at (0.505, 0.228) {\tiny{\citep{shortcuts-to-automata-transformer,co-occurence-isnot-factual-association-llm,shortcut-learning-foundation-lvm,spurious-features-memorised-dnn}}};

        \node[anchor=west] at (0.505, 0.109) {\tiny{\citep{prism-llm,panacea-pareto-llm,maxminrlhf-llm,pluralistic-position-rl,social-choice-alignment-rl}}};

        \end{scope}
        
    \end{tikzpicture}
    \caption{Hallucinations root causes from loss and optimisation. The \faCameraRetro\space and \faFile*[regular] icons indicate discussed modalities.}
    \label{fig:optimisation-root-cause}
    \Description[Root Cause Figure]{Root causes of hallucinations tracing back to loss and optimisation}
\end{figure}
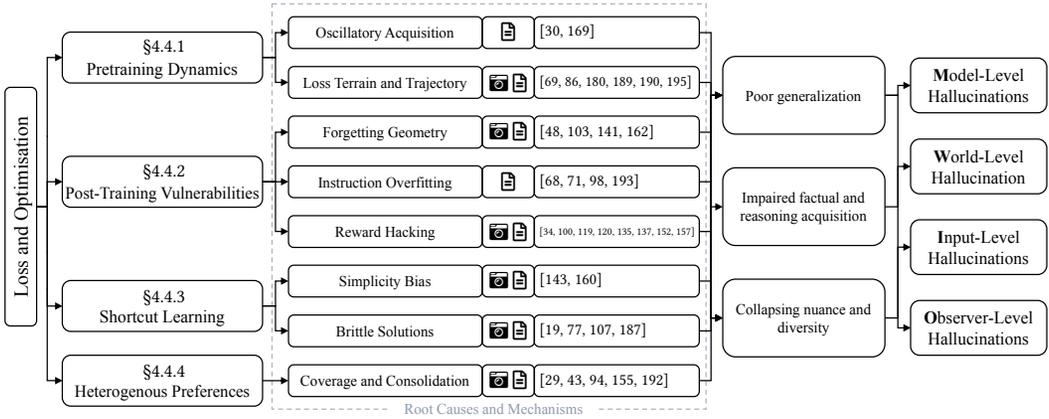

\subsubsection{Pretraining Dynamics}
\label{subsubsec:pretraining-dynamics}


Pretraining is the foundational phase in LLMs, LVLMs and TVMs. Here, models learn to optimise mostly self-supervised loss functions over a dataset. While often treated as a black box, emerging research has begun to reveal how fine-grained details within the optimisation process can affect final performance and generalisation. This section surveys key findings within the loss landscape and learning trajectories, highlighting how these factors both enable and hinder the emergence of desired capabilities.

\citet{how-llms-acquire-factual-knowledge-pretraining} found that during pretraining, the language log likelihood of factual knowledge follows an upward sawtooth pattern, reflecting cyclic phases of acquisition and partial forgetting. Duplicated facts amplify these oscillations, while larger models improve overall growth. This indicates that rare facts presented at infrequent intervals struggle to achieve sufficient likelihood maximisation for reliable decoding, regardless of training duration. To improve factual learning, the authors recommend outpacing the acquire-forget cycle by increasing sample diversity and batch size. \citet{initialisation-crucial-to-composition-llm} analysed how compositional reasoning in transformers emerges during pretraining. They found that small initialisation scales promote inferential learning, whose solutions capture highly organised features and compositional primitives. In contrast, larger scales favour symmetric learning, leading to disorganised features and pattern memorisation. \citet{reversal-curse-theoretical-understanding-llm} provided an optimisation perspective on directionality bias in LLMs (see Section \ref{subsubsec:directional-asymmetries}). They showed that unconstrained cross entropy loss optimisation hinders the learning of bidirectional associations. Given a logically equivalent relationship $A\leftrightarrow B$, weight updates for $A\rightarrow B$ do not necessarily update the reverse relation. \citet{pretraining-dynamics-icl-task-reg-learn} identified two core learning mechanisms in ICL for LLMs. First, models use pretraining knowledge to recognise tasks from examples. Second, models learn new tasks from given examples. Both learning mechanisms compete during pretraining: as one ability improves, the other often declines. This competition negatively affects overall ICL performance, leading to fluctuations rather than consistent growth. The authors show that a structured learning curriculum can mitigate the oscillatory effects of this competition. \citet{vit-training-dynamics} formulated generalisation conditions for vision transformers in the overfitting regime. By analysing the evolution of multi-head attention weights during pretraining, they proposed a precise mathematical criterion for predicting benign overfitting. \citet{autoregressive-next-token-icl-generalisation} presented rigorous PAC-Bayesian generalisation bounds for ICL in LLMs by accounting for the pretraining optimisation trajectory and duration. Specifically, they recommend warm starting and faster convergence for better generalisation. \citet{lipschitz-singularity-diffusion} demonstrated that TVMs, particularly when training with noise-prediction or velocity-prediction objectives, manifest Lipschitz singularities near the initial timestep as the partial derivative of the noise prediction network tends to infinity. Such behaviour is highly problematic and important to mitigate because it can lead to instability and potential errors in approximation and learning. \citet{parameter-symmetry-lvm} established theoretical guarantees demonstrating how loss-invariant transformations within the parameter space of deep neural networks can accelerate the convergence of SGD. Specifically, the ability to pivot within loss-invariant level sets enables the selection of parameter-symmetric points exhibiting higher local curvature, as indicated by the Hessian, which are correlated with improved generalisation.   

These studies show how unconstrained optimisation can hinder the effective acquisition of knowledge and generalisable skills. This implies that hallucinations stemming from pretraining dynamics can stem from various root causes such as the acquire-forget cycle, poor initialisation scales, oscillatory ICL acquisition, and non-conformance to specific generalisation bounds, trajectories and conditions. Targetted interventions of fine-grained mechanics within the optimisation process is therefore crucial in both improving general performance and reducing hallucinations.

\subsubsection{Post-Training Vulnerabilities}
\label{subsubsec:post-training-vulnerabilities}
Post-training refers to the process by which a raw pretrained model is transformed into usable, task-oriented systems. This stage usually involves three key techniques: instruction tuning, reinforcement learning from human feedback (RLHF), and domain finetuning. Unlike pretraining, which imparts general textual or image capabilities, post-training sharpens these general capabilities into more directed, goal-driven behaviours. However, this stage may also introduce new vulnerabilities and pathological behaviours. The section here examines how hallucinations may trace its roots back to specific mechanisms within each of these three post-training strategies.

Domain finetuning applies gradient updates to adapt a model to specialised fields. A key challenge here is catastrophic forgetting: the loss of general capabilities and previously acquired knowledge \citep{cata-forget}. While this issue is a visible and direct failure mode, we identify deeper root mechanisms within the loss landscape that drive forgetting and, ultimately, contribute to hallucinations. \citet{forget-peft-llm} analysed the loss landscape between two sets of LoRA parameters finetuned on consecutive tasks and identified a parametric valley path, a form of mode connectivity, linking the local minima of both tasks. This suggests that forgetting can be modulated via linear interpolation between task-specific parameters in the loss landscape. Using this interpolative approach, the authors successfully reduced factual hallucinations on domain-specific benchmarks. Both \citet{forget-loss-sharpness-llm} and \citet{forget-biased-llm} found that forgetting worsens as the loss landscape becomes sharper. Sharp curvatures result in large loss deltas with small parameter updates, an issue prominent when finetuning on novel datasets. Their interventions to flatten the loss landscape effectively reduced hallucinations on both new and previously learned tasks. Extending this, \citet{forget-linear-regresion-dnn} theoretically showed that learning tasks with larger eigenvalues, reflecting higher data variance and requiring larger updates, later in training worsens forgetting, especially with larger step sizes in high-dimensional spaces. These works indicate that trajectory geometries within the loss landscape during domain finetuning are a root cause of hallucinations. By controlling how aggressively models traverse the optimisation terrain, developers may reduce hallucinations stemming from catastrophic forgetting.

Instruction tuning involves training the model to follow textual commands. One major issue in this process is overfitting \citep{shi2024instruction-overfit}. \citet{it-limits-llm} analysed the output distributions of instruction-tuned LLMs and found that models mimicked the verbosity of training samples, producing overly detailed responses without sufficient factual grounding, and also tended to recall phrases verbatim from the instruction data to match prompt topics. Both \citet{lima} and \citet{model-really-follow-instructions} proposed that models often latch on to superficial cues, like format and style, during instruction tuning. LLMs trained on simplified or even semantically meaningless instruction samples can perform comparably to those trained on original ones. Aside from overfitting, \citet{context-parametric-inversion-instruct-tuning} identified a paradoxical phenomenon: instruction-tuned models, despite strong benchmark performance, often learn to ignore context. This issue stems from two types of instruction samples: those where context is necessary for correct answers, and those that closely match pretrained sequence structures. Early in training, context-dependent samples drive learning to promote context reliance. However, over time, gradients from pretrain-overlap samples dominate, gradually shifting the model away from context reliance and towards simply formatting outputs based on its pretrained knowledge. These studies reveal root mechanisms behind how instruction tuning can lead to hallucinations: models overfit to stylistic patterns and exploit overlaps with their pretraining data. This could result in verbose outputs that sound plausible and faithful, but contain factual and context hallucinations. Carefully curated data and optimisation regimes are crucial to mitigating this root mechanism.

RLHF is a powerful technique for aligning models with human preferences. A commonly used algorithm in RLHF is Proximal Policy Optimisation (PPO), which relies on a reward function to approximate human preferences and guide the model toward preferred outputs. However, PPO is susceptible to reward hacking, a phenomenon where models learn to exploit the reward function with idiosyncratic outputs, at the expense of quality. \citet{reward-hacking-rl} provided a rigorous theoretical framework for understanding reward hacking, showing that preventing it requires imposing strict constraints on both the policy set and the optimisation process. Supporting this, both \citet{new-definition-reward-hacking} and \citet{constrained-rlhf-reward-hacking} demonstrated that reward hacking can stem from aggressive optimisation and proposed constraints to bound learning. Additionally, \citet{inform} revealed that the root of reward hacking lies in the process of training the reward function itself, usually a neural network, which is prone to shortcut learning. \citet{sail-into-the-headwind} offered a data-centric explanation for reward hacking. They show that under-represented preference samples can lead to high-variance estimates, potentially generating excessive reward signals, even when the quality is bad. Taking a fresh perspective, \citet{odin-reward-shaping} points to a deeper root cause of reward hacking: human annotators are cognitively biased toward verbose and complex-sounding responses, regardless of factuality or quality. This biased preference is thus inherited by the reward model during training. \citet{purm} theorised that the standard Bradley-Terry reward model used in PPO inherently lacks support for expressing uncertainty, which leads to extreme and overconfident signals that are susceptible to hacking. There exist reward-free RLHF methods \citep{dpo}, but even those have been observed to exhibit over-optimisation trends similar to reward hacking \citep{daa-reward-overoptimisation-llm}. While reward hacking presents a straightforward pathway to hallucinations, our investigation reveals deeper factors that underpin reward hacking, specifically data availability, human biases, optimisation dynamics, and reward function design. These root causes and mechanisms ultimately drive low-quality, idiosyncratic, and potentially hallucinatory outputs.

Taken together, these findings indicate how specific mechanisms within each of these three post-training strategies can serve as root causes of hallucinations. In domain finetuning, sharp loss landscapes and interference between tasks drive catastrophic forgetting. In instruction tuning, stylistic artifacts push models toward instruct overfitting. In RLHF, aggressive optimisation, preference sparsity and annotation biases encourage reward hacking. In turn, catastrophic forgetting, instruction overfitting and reward hacking lead to seemingly well-structured outputs that are factually ungrounded, contextually unfaithful and idiosyncratic, which ultimately results in factual, context, and user-perceived hallucinations.

\subsubsection{Shortcut Learning}
\label{subsubsec:shortcut-learning}
Research has shown that LLMs and LVLMs, despite being able to solve seemingly complex tasks, often tend to learn "easy" solutions over robust abstractions during training \citep{simplicitiy-bias-dnn,shortcut-learning-dnn}. \citet{simplicity-bias-2layernn} demonstrated that neural networks exhibit a simplicity bias during learning. Analysis reveals that features consistently cluster around limited directions, extrema of a simpler data-dependent function, regardless of layer width or dataset complexity. In transformers, \citet{simplicity-bias-transformer-nlp} showed, by synthesising and controlling degrees of interaction, that LLMs exhibit a simplicity bias. Models prioritise learning simpler patterns (like bigrams) first before learning more complex ones as training progresses. Having a simplicity bias may be beneficial in some cases \citep{simplicity-bias-good,simplicity-bias-good-2}. However, it becomes concerning when these biases turn into shortcut learning \citep{simplicity-bias-shortcut-1,simplicity-bias-shortcut-2}. \citet{shortcuts-to-automata-transformer} demonstrated shortcut learning in transformers by training it to simulate rule-based, state-transition machines (semiautomatons). Here, a shortcut solution refers to a model using fewer layers than the sequence length to simulate automaton behaviour. Transformers trained with SGD consistently learn shortcut solutions to all automata with depth logarithmic to sequence length by leveraging algebraic structures. These shortcut solutions are brittle, failing to generalise to unseen sequence lengths. \citet{co-occurence-isnot-factual-association-llm} observed, using knowledge triplets, that LLMs trained on directly stated facts learned more shortcut co-occurrence statistics, which did not generalise to complex questions. \citet{shortcut-learning-foundation-lvm} proposed that shortcut learning is primarily driven by how easily a feature is extracted, rather than solely by a feature's correlation with class labels. On image datasets, vision models tend to learn readily available but statistically suboptimal features. For example, prominent backgrounds or textures. Supporting this, \citet{spurious-features-memorised-dnn} mathematically demonstrated that spurious features can be learnt even if statistically independent from the true label. These findings here show how various problem-specific drivers behind shortcut learning can serve as root causes of hallucinations. Mitigation strategies will need to be tailored and targetted, such as manipulating backgrounds in images, identifying known theoretical shortcuts, or suppressing simplicity biases. Since shortcut learning is often obscured by in-distribution benchmarks, another possible solution is with more robust LLM and LVLM evaluations, though they are fraught with their own set of challenges (See section \ref{subsec:misleading-evaluations}). 

\subsubsection{Heterogeneous Preferences}
\label{subsubsec:heterogeneous-preferences}
Optimising LLMs, LVLMs, and TVMs for human alignment is fundamentally complicated by the heterogeneity of human preferences, even in tasks that appear neutral. For example, scientific questions involving emerging terminologies or controversial procedures often elicit divergent expectations across users. In such cases, a single output risks alienating parts of global audiences. \citet{prism-llm} collected preference data from 75 countries, revealing deep disagreements in how users interpret model responses on value-laden issues. Similarly, \citet{panacea-pareto-llm} critiqued scalar alignment labels for collapsing diverse and nuanced preferences into a single, overly simplistic objective. This reductive approach risks marginalising under-represented needs in subjectively complex tasks. From a theoretical standpoint, \citet{maxminrlhf-llm} showed that single-reward RLHF is mathematically incapable of capturing the full spectrum of sub-population preferences, while \citet{pluralistic-position-rl} argued that failure to embrace pluralism may result in algorithmic monocultures that amplify social inequities. \citet{social-choice-alignment-rl} further warned that ad hoc aggregation of divergent views can marginalise stakeholders and exacerbate tensions. Models optimised under narrow or homogenised preference regimes risk producing outputs that reflect implicit biases while failing to accommodate dissenting perspectives and expectations. While these failures are not classically defined as hallucinations in academic contexts, real-world users often perceive these outputs as incoherent, untrue, or nonsensical in practical applications \citep{hci-user-intent-llm-1,hci-user-intent-llm-2}. Addressing these user-perceived hallucinations requires new alignment strategies capable of navigating conflicting and pluralistic human preferences.

\subsection{Misleading Evaluations}
\label{subsec:misleading-evaluations}

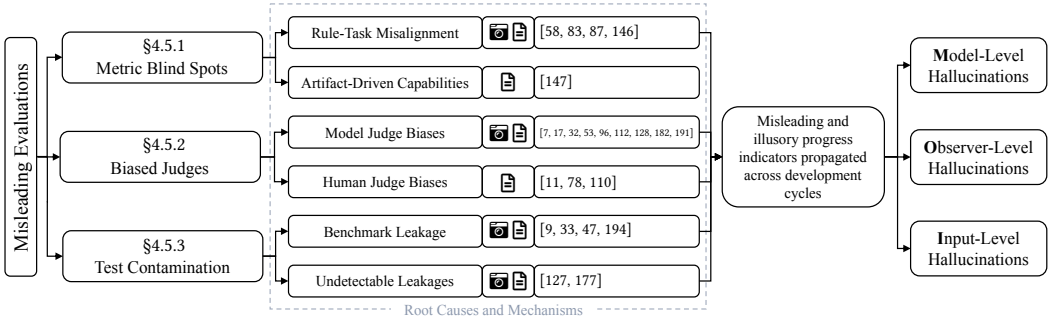
\begin{figure}
    \centering
    \begin{tikzpicture}
        \node[anchor=south west, inner sep=0] (image) at (0,0) {\includegraphics[width=\textwidth]{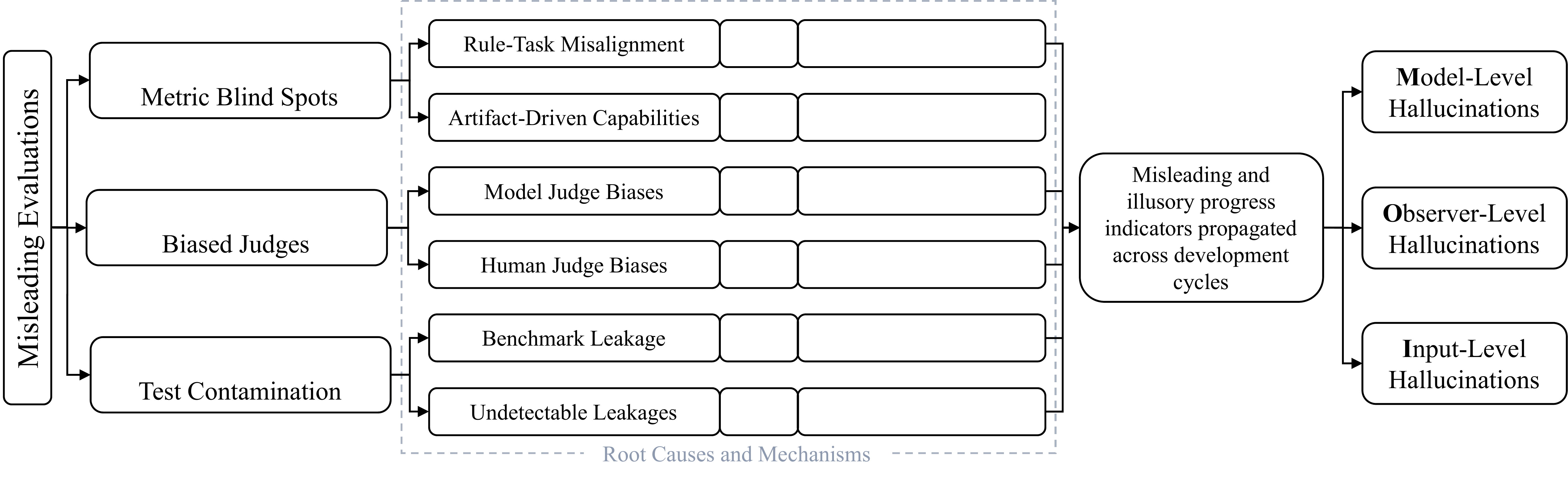}};

        \begin{scope}[x={(image.south east)}, y={(image.north west)}]

        \node[anchor=center] at (0.155, 0.865) {\scriptsize{\S\ref{subsubsec:metric-blind-spots}}};
        \node[anchor=center] at (0.155, 0.56) {\scriptsize{\S\ref{subsubsec:biased-judges}}};
        \node[anchor=center] at (0.155, 0.25) {\scriptsize{\S\ref{subsubsec:test-contamination}}};

        \node[anchor=center] at (0.484, 0.91) {\scriptsize{\faCameraRetro\space\faFile*[regular]}};
        \node[anchor=center] at (0.484, 0.755) {\scriptsize{\faFile*[regular]}};
        
        \node[anchor=center] at (0.484, 0.6) {\scriptsize{\faCameraRetro\space\faFile*[regular]}};

        \node[anchor=center] at (0.484, 0.445) {\scriptsize{\faFile*[regular]}};

        \node[anchor=center] at (0.484, 0.3) {\scriptsize{\faCameraRetro\space\faFile*[regular]}};

        \node[anchor=center] at (0.484, 0.145) {\scriptsize{\faCameraRetro\space\faFile*[regular]}};

        \node[anchor=west] at (0.505, 0.91) {\tiny{\citep{fid-limitations,perplexity-curse,perplexity-long-context-llm,perplexity-curse-knowledge-extraction-llm}}};

        \node[anchor=west] at (0.505, 0.755) {\tiny{\citep{emergent-mirage-llm}}};

        \node[anchor=west] at (0.505, 0.6) {\scalebox{.41}{\citep{length-controlled-alpacaeval-llm,mtbench-judge-chatbot-arena-llm,visit-bench-lvlm,lmexam-llm,narcissistic-evaluators-llm,evaluator-self-preference-llm,benchmarking-cognitive-biases-llm,calm-llm,llm-judge-bias-study-llm}}};

        \node[anchor=west] at (0.505, 0.445) {\tiny{\citep{re-eval-elo-eval,human-feedback-not-gold-standard,preference-disagreement-llm}}};

        \node[anchor=west] at (0.505, 0.3) {\tiny{\citep{leak-cheat-repeat,data-leakage-swe-llm,ts-guessing-llm, right-way-eval-testleak-lvlm}}};

        \node[anchor=west] at (0.505, 0.145) {\tiny{\citep{proving-test-contamination-llm,benchmarking-benchmarks-llm}}};

        \end{scope}
        \end{tikzpicture}
    \caption{Hallucinations root causes from misleading evaluations. The \faCameraRetro\space and \faFile*[regular] icons indicate discussed modalities.}
    \label{fig:eval-root-cause}
    \Description[Root Cause Figure]{Root causes of hallucinations tracing back to misleading evaluations}
\end{figure}

\subsubsection{Metric Blind Spots}
\label{subsubsec:metric-blind-spots}
Evaluation metrics play a central role in shaping how developers improve generative models. Yet, many widely adopted metrics, originally designed for narrow, well-controlled benchmarks, struggle to capture the complexity of modern LLM and TVM outputs. These metrics often misalign with real task performance and may hinder efforts in identifying and mitigating failure modes. \citet{fid-limitations} challenged the reliability of the Fréchet Inception Distance (FID) score as the standard metric for assessing the quality of images produced by TVMs. FID, trained on a narrow set of ImageNet classes, cannot effectively capture the rich and varied outputs of modern image generators. In experiments, FID scores exhibited significant misalignment between FID and human perceptual judgment. Furthermore, the authors empirically invalidated FID's core operational assumption, showing that image features are not normally distributed. Perplexity is a widely used evaluation metric in language modelling benchmarks to indicate general performance. However, studies have found it to be a misleading proxy. \citet{perplexity-curse}, \citet{perplexity-long-context-llm} and \citet{perplexity-curse-knowledge-extraction-llm} found that models scoring near-perfect perplexities in tasks requiring long-context information extraction, perform poorly on similar test settings, especially when facts in test documents are interleaved or appear later. They argue that perplexity, focused on autoregressive next-token prediction, incentivises rigid memorisation of facts within fixed context patterns over generalisable retrieval. Thus, evaluating for low perplexities alone is a misleading approach to preventing hallucinations in real-world retrieval and reasoning tasks. In these settings, relevant information is often fragmented across long contexts and embedded within complex, non-linear narratives. Perplexity is not the only metric under scrutiny; concerns have emerged about how choice of metrics affect our interpretation of model properties. \citet{emergent-mirage-llm} argued that emergent capabilities in LLMs are not intrinsic properties, but rather artifacts of evaluation metrics that scale non-linearly or discontinuously with per-token error rates. They show that insufficient metric resolution (defined as 1/test dataset size) and sampling in the large-parameter regime can result in sharp, unpredictable changes in test performance. Analysis reveals that emergent abilities mostly appear under specific non-linear metrics, and vanish under linearised, continuous variants with sufficient resolution. Together, these findings reflect a broader concern that commonly relied upon metrics may distort our understanding of model limitations. Addressing hallucination will require more principled approaches to metric selection, ones that accurately reflect the generalisation and reliability we expect in real-world settings.

\subsubsection{Biased Judges}
\label{subsubsec:biased-judges}
LVLMs and LLMs are now increasingly evaluated on elaborate, open-ended tasks to judge their real-world applicability and guide future development iterations. Traditional rule-based metrics cannot capture the complex outputs required in such cases. In response, the community has adopted two main evaluation approaches: model-as-a-judge, and human judges. Both are crucial for guiding post-development efforts in model iterations. However, each introduces distinct biases that can distort perceived capabilities and mask crucial flaws during evaluation. Model-as-a-judge uses a powerful LVLM or LLM to cheaply and quickly assess the outputs of other models. However, there are five serious biases to consider. Some have demonstrated that model judges systematically favoured longer responses, regardless of content \citep{length-controlled-alpacaeval-llm, mtbench-judge-chatbot-arena-llm}. Their win rates could be flipped, as high as 90\% of the time, by crafting low-quality but verbose responses. Additionally, model judges also tend to consistently rate outputs generated by the same model family higher than other models or humans, regardless of quality \citep{visit-bench-lvlm, lmexam-llm, narcissistic-evaluators-llm, evaluator-self-preference-llm}. More troubling, scores from model judges could change by as much as 80\% by simply changing the positional order of responses \citep{mtbench-judge-chatbot-arena-llm, benchmarking-cognitive-biases-llm, llms-are-unfair-evaluators-llm}. Furthermore, studies showed that model judges are unduly influenced by superficial markers of authority and popularity, such as fabricated citations and statistics, in deciding output scores \citep{calm-llm, llm-judge-bias-study-llm, benchmarking-cognitive-biases-llm}. Given the biases introduced by the model-as-a-judge approach, one might turn to human judges, often seen as costlier but more reliable alternatives. Methods like preference feedback and Elo-rating arenas are common, yet studies show they introduce their own issues. \citet{re-eval-elo-eval} revealed that Elo-rating arenas can be unintentionally or strategically manipulated through prompt redundancy and specialisation. Simulations revealed that Elo ratings reward overrepresented skills in the prompt distribution, which limits their ability to assess balanced skill development. \citet{human-feedback-not-gold-standard} showed that human annotators were prone to cognitive and perceptual biases. Annotators tended to overlook factual inaccuracies in responses that appear assertive or complex. They also frequently conflate distinct quality dimensions. For example, a response rated poorly for helpfulness is more likely to be penalised in unrelated areas, like factuality, even when accurate. \citet{preference-disagreement-llm} pointed out that feedback collection methods, ratings and rankings, can introduce biases. About 60\% of preferences acquired from ratings contradict those collected from rankings. Human annotators rated verbose responses higher yet preferred concise ones in pairwise rankings. Models trained on rankings outperformed those trained with ratings during ranking evaluations, and vice versa. These findings reveal that when evaluating LLMs and LVLMs on elaborate, open-ended tasks, both model-based and human judging approaches are prone to systemic biases. Model judges, while scalable and efficient, exhibit preferences for verbosity, self-similarity, positional, and authority artifacts. Human judges, often considered the gold standard, are similarly influenced by cognitive heuristics, evaluation design, and ambiguous quality dimensions. These evaluation signals, central to guiding model development, risk being distorted by these biases. A root cause of hallucination may be that, over successive evaluation-development cycles, flawed outputs were either left undetected, or worse, reinforced by biased evaluation signals.

\subsubsection{Test Contamination}
\label{subsubsec:test-contamination}
Test contamination in LLMs refers to the unintended presence of test data within a model’s training set. Traditionally, strict separation between training and test data helps ensure that models learn meaningful and generalisable connections to perform effectively in the real world. For LLMs trained on massive, publicly scraped datasets, enforcing this separation is increasingly unrealistic. Studies have shown that popular benchmarks like MMLU and ARC contain samples overlapping with pretraining data \citep{leak-cheat-repeat, data-leakage-swe-llm, ts-guessing-llm}. This problem also affects the multimodal domain. LVLMs inherit data leaks from their underlying LLMs, enabling them to solve image-text problems by memorizing text pattern alone without visual inputs \citep{right-way-eval-testleak-lvlm}. Despite the various techniques used to detect test contamination, the sheer scale and unstructured nature of pretraining datasets make it implausible to filter out all leakages \citep{proving-test-contamination-llm, benchmarking-benchmarks-llm}. The presence of test contamination undermines the credibility of benchmark evaluations; performance gains may be illusory, while critical flaws remain hidden. One root cause of persistent hallucinations may stem from models being deployed under a false sense of progress. Ensuring cleaner benchmarks is therefore not just an academic pursuit, but a crucial step for establishing real improvements in reliability.

\section{Discussion}
\label{sec:discussion}
This survey aims to establish a principled, unified, and modality-agnostic framework for understanding hallucinations in LLMs, LVLMs, and TVMs. It begins by proposing a general formal definition of hallucination that is not tied to specific tasks or output modalities, but instead grounded in fundamental modelling principles to ensure broad applicability. Following that, the survey investigates the root causes of hallucination by tracing them to identifiable mechanisms across five key stages of a model’s lifecycle. These findings are presented in two complementary ways: by uncovering common patterns and shared causes across all three model types, and by exploring the unique challenges and failure modes specific to each of them.

The Model–Observer–World–Input (MOWI) framework provides four structured levels for defining hallucinations across all three model types. Each level can be mapped to concrete causes within a model's lifecycle. Model-level hallucinations arise from failures in density estimation, such as erroneous interpolation or extrapolation of the data manifold. These errors stem from low-quality training data, sporadic breakdowns in architectural mechanisms, optimisation dynamics, and degenerate evaluation cycles. Observer-level hallucinations occur when model outputs diverge from human expectations, even when technically valid. Users perceive outputs as nonsensical or incoherent when they lack nuance, appear idiosyncratic, or fail to fulfil expected tasks. These failures are typically rooted in ad-hoc homogenisation, exclusion of diverse perspectives, reward hacking, and inaccurate extrapolation from in-context examples. World-level hallucinations reflect epistemic and aleatoric limitations: what the model cannot know due to gaps or randomness in the external world. These are driven by systemic omissions in training data, ambiguous contexts, and the way knowledge is preferentially encoded, acquired, and sequentially generated. Input-level hallucinations emerge when models are forced to operate when conditioned on unreliable or adversarial contexts. Here, the quality of the input, whether user-provided, self-generated, or agentic, is critical. Models often struggle with meta-reasoning and are further constrained by structural biases inherited from their training regime and evaluators. Together, the MOWI framework offers a unified approach for diagnosing and categorising hallucinations across real-world LLM, LVLM and TVM AI systems.

Four broad themes emerge from the survey's root cause investigation, revealing how hallucinations manifest as structural and predictable outcomes of how these models are trained and used. First, models hallucinate when pushed outside their training distribution. For instance, while models may perform reliably on basic tasks such as counting, their performance deteriorates on variations like fractional counting (e.g., drawing pizzas), algorithmic shifts, or larger numerical ranges, due to insufficient training data and architectural support. More broadly, this indicates an important caveat: LLMs, LVLMs and TVMs are strong generalist agents insofar as "novel" tasks remain within the structure of what they’ve seen. When tasks fall into low-data regimes within the training distribution, whether due to counterfactual conditions, increased compositional complexity, or under-represented scene types, models are more likely to fail. Second, models hallucinate along predictable axes of inherited biases. They internalise directional tendencies present in language, such as conversational flow and mathematical procedures. During RLHF, reward signals often misalign with factuality or coherence. Human raters tend to rely on superficial heuristics, favouring outputs that merely appear "good". Annotators drawn from a narrow group risk marginalising irreconcilable differences in contested domains, such as emerging scientific theories or debatable medical practices. Moreover, self-dialogues or agentic debates can reinforce shared misconceptions. Collectively, these examples suggest that hallucinations are not just errors, but predictable byproducts of biases embedded in their data, incentives, and interactions. Third, fine-grained internal dynamics within both architectural and optimisation stages, commonly overlooked or abstracted away, may define the boundary between robust generalisation and systemic hallucinations. Transformer-based models and denoising diffusion architectures possess implicit inductive biases that lead to preferred manners of learning, which may diverge from intended task objectives. During training, the optimisation trajectory, loss terrain, and initialisation strategy influences performance and generalisation. Hallucinations may not be mere aberrations, but foreseeable consequences of architectural predispositions and optimisation processes. Fourth, evaluation reform is essential for progress. Rule-based metrics, such as perplexity or FID, cannot fully capture the multifaceted outcomes developers aim to optimise for. More concerning is the increasing popularity of model judges, which are demonstrably biased. Even human evaluators, often considered the gold standard, are biased towards superficial presentation and framing effects. In Elo arenas, models can score higher by specialising on overrepresented capabilities. Traditional train-test splits, intended to probe generalisation, are now increasingly difficult to enforce. Without robust evaluations, developers risk perpetuating and exacerbating hallucinatory tendencies in models. These four broad themes taken together indicate that hallucinations are not incidental anomalies but systematic and predictable artifacts rooted in the ways models are trained and used.

\section{Future Directions}
\label{sec:future-directions}
Building on the themes discussed, several promising avenues for mitigating hallucinations in machine learning systems emerge. Models tend to hallucinate when pushed beyond their training data, particularly in rare tasks and expert domains. Machine teaching offers a promising solution by empowering domain experts directly through user-friendly tools and organised pipelines \citep{machine-teaching, machine-teaching-2}. By decoupling algorithm design from model building, development on a wide range of rare and expert tasks can be made more scalable, accessible, and maintainable. Additionally, test-time adaptation offers another complementary strategy by enabling models to update their internal representations dynamically during inference, using self-supervised signals from new inputs \citep{tta-vlm1,tta-2,tta-3}. Together, these approaches aim to ensure that models remain grounded within their intended data distributions, even as real-world conditions shift.

Another future direction stems from previously discussed insights on how the internal learning dynamics of optimisation and architectural choices can influence robust learning. Rather than relying solely on theoretical efforts, emerging efforts in mechanistic interpretability seek to uncover the computational anatomy of intelligence within neural networks, tracing how specific neuron pathways evolve to represent meaningful abstractions \citep{spatial-reasoning-vlm-hard, circuit-analyses-1, circuit-analyses-2}. However, this work raises deeper questions about the very nature of abstractions. Future research could benefit from formalising how abstractions arise and are structured, both in human and machine cognition \citep{arc-abstraction-1,abstraction-2,abstraction-3}. Moreover, for tasks where abstractions within parametric neural networks prove brittle or opaque, hybrid approaches that integrate symbolic reasoning with deep learning could offer a more stable and interpretable foundation for generalisation and reasoning \citep{neurosymbolic-1,neurosymbolic-2,neurosymbolic-3}.

Finally, advancing the evaluation of hallucinations remains a crucial frontier. As our discussions highlight, traditional evaluation practices can systematically overlook failure modes. A more proactive approach could involve adopting red-teaming strategies: systematically stress-testing models under adversarial and realistic conditions to expose vulnerabilities \citep{red-team-1,red-team-2,red-team-3}. Beyond identifying isolated errors, there is value in emerging theoretical efforts reframing hallucinations not as incremental faults, but as part of a Pareto trade-off \citep{gen-restorative-models-info-theory-hallucination}. Embracing this perspective encourages a more nuanced exploration of task-specific tolerances for hallucinations \citep{hallu-tradeoff-1,hallu-tradeoff-2}. Crucially, future evaluations must also account for observer-perceived hallucinations: outputs that, while technically defensible, may appear incorrect to users based on their local environments and lived experiences. Social Choice Theory offers a promising framework for integrating diverse user preferences and epistemic standards into more inclusive definitions of model reliability \citep{social-choice-alignment-rl, social-choice-1}.

While this survey offers a foundation for understanding hallucinations, a few key limitations remain. Current research has largely focused on textual language models, with vision-language models often interpreted through textual anchors; future work should more deeply explore the visual and multimodal dimensions of hallucination. Additionally, less common modalities, such as audio, demand greater attention. Deeper theoretical analysis, stronger mechanistic understanding, and more systematic frameworks for discussion are also needed. Finally, the landscape presented here is not exhaustive; many perspectives and nuances remain unexplored, offering rich opportunities for future investigation.

\section{Conclusion}
\label{sec:conclusion}
In this survey, we systematically traced the root causes and mechanisms behind hallucinations in Large Language Models (LLMs), Large Vision-Language Models (LVLMs), and Text-to-Image Vision Models (TVMs) across their full lifecycle: from data, architecture, inference, loss optimisation to evaluation. By proposing a unified, modality-agnostic definition of hallucinations and identifying shared vulnerabilities, we aim to bridge fragmented research efforts and provide a more unified understanding of these failures. Through this comprehensive survey, we gained critical insights revealing that hallucinations are not isolated or sporadic errors, but rather predictable and principled behaviours rooted in design choices, training dynamics, and deployment practices. These insights suggest that mitigating hallucinations demands addressing challenges across multiple frontiers, including data distribution adaptation, mechanistic interpretability, abstraction learning, and the development of new evaluation paradigms. As these three types of generative models continue to scale and permeate critical real-world domains, our findings highlight the urgent need for principled and unified strategies to ensure the reliability of AI systems.

\bibliographystyle{ACM-Reference-Format}
\bibliography{main}

\appendix
\end{document}